\documentclass{article} 
\usepackage{iclr2027_conference,times}

\usepackage{amsmath,amsfonts,bm}

\def\eqref#1{equation~\ref{#1}}

\def\1{\bm{1}}

\DeclareMathAlphabet{\mathsfit}{\encodingdefault}{\sfdefault}{m}{sl}
\SetMathAlphabet{\mathsfit}{bold}{\encodingdefault}{\sfdefault}{bx}{n}

\usepackage{hyperref}
\usepackage{url}
\usepackage{xcolor}
\usepackage{graphicx}
\usepackage{amsmath,amssymb}
\usepackage{placeins}
\usepackage{booktabs}
\usepackage{algorithm}
\usepackage{algpseudocode}

\title{Learning Computational Structure\\ for Sample-Efficient Prediction}

\author{
\textbf{Hangyu Zhou}$^{1}$ \qquad
\textbf{Constantine Dovrolis}$^{2}$ \qquad
\textbf{Stephen Mussmann}$^{1}$
\\[0.8em]
\begin{tabular}{@{}l@{}}
$^{1}$Georgia Institute of Technology \\
$^{2}$University of Cyprus
\end{tabular}
}

\iclrfinalcopy 
\begin{document}

\maketitle
\begingroup
\renewcommand{\thefootnote}{}
\footnotetext{Correspondence to
\texttt{hzhou411@gatech.edu}}
\endgroup

\fancyhead[L]{Preprint}

\begin{abstract}
Statistical learning in high dimensions suffers from the curse of dimensionality.
Fortunately, many functions of practical interest are compositionally sparse,
meaning they admit a composition of local functions, each depending on only a
few variables, allowing sample complexity to depend on local arity rather than
ambient dimension.
Yet realizing this benefit requires exploiting hidden compositional structure,
which theory does not establish for practical gradient-based training.
We study this gap with CSFRBench, a controlled benchmark that hides the target's
compositional computation during training but retains it for diagnosis, and find
that dense networks and sparsity-based baselines leave much of that benefit unrealized.
Inspired by the spatial organization of biological neural circuits,
we introduce \emph{Learned Wiring} (LeWi): each neuron has a learned
low-dimensional embedding, only nearby neurons connect, and embeddings and
weights are trained jointly.
LeWi improves sample efficiency on targets with hierarchical structure, intermediate reuse, or independent modules. 
Across 32 additional targets generated by independently sampled computation
graphs, LeWi is the most sample-efficient method not given the
target graph, with 27\% lower mean learning-curve area than dense
networks. Dense networks perform better on targets with a single
high-arity function.
LeWi's representations and connectivity align with the target computation.
On 17 world-model prediction tasks spanning control, navigation, and manipulation,
LeWi beats the dense baseline on 16 and wins 99/119 baseline comparisons.
These results show that jointly learning connectivity and weights can exploit
computational structure to improve sample efficiency.
\end{abstract}
\begin{figure}[h] \centering \includegraphics[width=\linewidth]{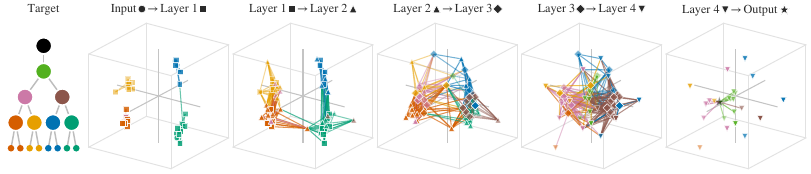} 
\vspace{-1.5em}
\caption[LeWi develops embedding geometry aligned with the target computation.]{ \textbf{LeWi develops embedding geometry aligned with the target computation.} Learned neuron embeddings across successive layers of LeWi trained on a depth-4 tree. Colors indicate post hoc associations with intermediate values in the target computation. Inputs unused by the target and hidden units without a target association are omitted. \href{https://zhou-hangyu.github.io/lewi/}
{Watch the geometry emerge during training}:
\protect\url{https://zhou-hangyu.github.io/lewi/}. } \label{fig:teaser} \end{figure}
\section{Introduction}

Learning in high dimensions is statistically difficult: for general smooth functions, sample complexity grows exponentially with ambient dimension~\citep{stone1982optimal}.
Yet many real-world learning problems are not arbitrary high-dimensional mappings. Their outputs arise from interactions among objects, mechanisms, subsystems, or latent variables, and may be expressed through reusable intermediate computations, each depending on only a few inputs (low arity).
Such decompositions exhibit \emph{compositional sparsity}~\citep{poggio2024compositional}. 
Under suitable smoothness assumptions, statistical theory shows that the sample complexity of regularized estimators grows exponentially with local arity rather than ambient input dimension~\citep{schmidt2020nonparametric}.

These statistical guarantees, however, assume global or near-global optimization and therefore do not establish comparable sample efficiency under practical gradient-based training. 
This raises our central question: \emph{can
gradient-based training leverage hidden compositional structure to improve
sample efficiency using only input--output examples?}

We study this question through compositionally sparse function regression
(CSFR), where the target is composed of low-arity local functions connected
by a sparse computation graph. We introduce CSFRBench, a controlled benchmark
that enables systematic evaluation of whether gradient-based learners can
exploit hidden compositional structure from input--output examples. 
CSFRBench varies computation graphs and
local-function families while retaining the target computation for diagnosis,
allowing predictive sample efficiency to be studied alongside learned internal
organization. 
Across its compositional settings, dense networks and sparsity-based
baselines, including methods that adapt sparsity at the level of individual
connections, remain substantially less sample-efficient than same-backbone
references whose fixed sparsity patterns encode target connectivity.
Existing gradient-based learners therefore leave much of the
sample-efficiency benefit associated with compositional structure unrealized.

Biological neural circuits provide one source of inspiration. 
Their connectivity is strongly shaped by spatial organization, and a long-standing line of work on \emph{wiring economy} studies how neural systems trade off communication efficiency with the cost of long-range connections, favoring structured local connectivity \citep{hiesinger2021self,cang2013developmental,chklovskii2004maps,bullmore2012economy}.
This suggests that connectivity may be more naturally parameterized through a shared spatial organization than by treating each connection independently.

Motivated by this principle, we introduce \emph{Learned Wiring} (LeWi),
which parameterizes network connectivity through learned neuron embeddings.
Each neuron has a low-dimensional embedding, and adjacent-layer neurons
connect when their embeddings lie within a fixed radius. The task loss jointly
learns the embeddings and connection weights: embeddings determine where
computation can flow, while weights determine what is computed. Because each
neuron's embedding participates in many connections, a single embedding update coordinates multiple
connectivity decisions.

LeWi improves sample efficiency on CSFRBench targets with hierarchical
structure, intermediate reuse, or independent modules.
Relative to dense networks, LeWi reduces mean learning-curve
area by 20\% on the depth-4 tree, 34\% on the graph with reuse,
and 56\% on flat modules, and by 27\% across 32 independently
sampled depth-4 graphs.
Dense networks perform better on the high-arity control, while LeWi's advantage increases across all settings
when nuisance inputs are added. These predictive gains are accompanied by
target-aligned internal organization. Target intermediates are more linearly
decodable from LeWi's hidden activations than from dense networks, and counterfactual
interventions show stronger agreement with the target
computation. LeWi's learned connectivity also aligns with target
dependencies, achieving more than three times the edge F1 of initialization
and randomized controls on average across four structured settings.

Beyond CSFRBench, across 17 world-model prediction tasks, LeWi improves over
the dense network baseline on 16 and wins 99 of 119 task--baseline comparisons,
with broad gains across control and navigation and more variable performance on
manipulation tasks. Together, these results show that learning connectivity
through shared neuron embeddings can help gradient-based training exploit
computational structure to improve sample efficiency.
\section{Compositionally sparse function regression}
\label{sec:csfr}

Let $G=(V,E)$ be a directed acyclic graph (DAG) representing the target computation, with input nodes
$x_1,\ldots,x_d$ corresponding to the input coordinates and an output
node $v_{\mathrm{out}}$.
For each input node, set $z_{x_j}=x_j$.
Each non-input node $v$ is associated with a local function $g_v$ and
computes
$z_v=g_v((z_u)_{u\in\mathrm{pa}_G(v)})$,
where $\mathrm{pa}_G(v)$ denotes the parents of $v$.
The resulting function is $f_G(x)=z_{v_{\mathrm{out}}}$.
A function is \emph{$k$-compositionally sparse} if it admits such a
representation with at most $\operatorname{poly}(d)$ non-input nodes and
$|\mathrm{pa}_G(v)|\leq k$ for every non-input node $v$, where $k$ is independent of $d$
\citep{poggio2024compositional,danhofer2025position}.
In compositionally sparse function regression (CSFR), we observe samples
$(\mathbf{x}_i,y_i)_{i=1}^n$ drawn from
$Y=f^\star(X)+\epsilon$, with $\mathbb{E}[\epsilon\mid X]=0$, where
$f^\star$ is $k$-compositionally sparse. 
The target computation graph is hidden from the learner and, because multiple
computation graphs can represent the same function, need not be identifiable
from input--output data. We therefore use it as a reference for structural
alignment rather than requiring graph recovery. Appendix~\ref{app:statistical-background}
summarizes corresponding statistical rates under smoothness assumptions.
\section{Related work}

\paragraph{Learning compositional functions.}
Classical nonparametric regression suffers from the curse of dimensionality
\citep{stone1982optimal}, whereas compositional function classes can admit rates
governed by local-function arity and smoothness
\citep{schmidt2020nonparametric}. Related statistical guarantees have been established for
sparse, norm-constrained, and fully connected neural networks
\citep{ohn2022nonconvex,huang2026learning,nagler2026optimal}.
Recent work argues that compositional sparsity may help deep networks avoid the
curse of dimensionality while identifying learnability and optimization as open
questions \citep{poggio2024compositional,danhofer2025position}.
A natural way to study such questions empirically is with controlled synthetic
targets, which have long been used to study nonlinear interactions and
sensitivity
\citep{friedman1991multivariate,ishigami1990importance,sobol1990sensitivity}.
More recently, \citet{cagnetta2024deep} study representation learning and sample
complexity using a fixed hierarchical classification model. CSFRBench instead
varies the target computation graph, hides it from the learner, and retains the
graph and intermediate values for evaluation, allowing predictive sample
efficiency and alignment with the target computation to be studied together.

\paragraph{Learned connectivity and neural wiring.}
Spatial and connection-cost biases have long been used in artificial neural
networks to encourage modular, topographic, or interpretable organization
\citep{jacobs1992computational,clune2013evolutionary,mengistu2016evolutionary,
blauch2022connectivity,achterberg2023spatially,liu2023seeing,deb2025toponets,zhang2025brain},
motivated in part by biological wiring principles
\citep{hiesinger2021self,cang2013developmental,chklovskii2004maps,
bullmore2012economy}.
These approaches impose an explicit spatial geometry and favor short or
low-cost connections.
Sparse-training methods instead adapt connectivity through learned gates or
masks, or through weight- and gradient-based topology updates
\citep{louizos2017learning,wortsman2019discovering,evci2020rigging,
jayakumar2020top,tang2022automatic,lasby2024dynamic}, while
\citet{lin2026compositional} infer dependencies from data to construct a fixed
sparse architecture before weight optimization.
Unlike these approaches, LeWi couples connectivity decisions through shared
neuron embeddings.
Closest to LeWi, \citet{karaletsos2018probabilistic} and
\citet{karaletsos2020hierarchical} use neuron embeddings to define structured
priors over Bayesian neural-network weights, while \citet{zhou2022permutation}
and \citet{erb2025training} generate weight values from relations between neuron
embeddings.
LeWi instead jointly learns neuron embeddings and connection weights, using
embedding distance to determine connectivity without an explicit wiring-cost
regularizer.

\paragraph{Modularity in neural networks.}
Prior work on modularity in trained neural networks includes graph-based
analyses of clusterability and local specialization
\citep{filan2021clusterability,hod2021quantifying}, masking and pruning methods
that identify subnetworks associated with specific subcomputations or
hierarchical task structure
\citep{csordas2020neural,lepori2023break,malakarjun2023neural}, and studies of
specialized neuron groups in pretrained Transformers
\citep{zhang2023emergent,qiu2024unlocking}.
These works characterize or extract modular organization from trained networks,
whereas LeWi parameterizes and learns connectivity for improving predictive sample
efficiency.
\section{LeWi: Learned Wiring}
\label{sec:methods}

We introduce LeWi (Learned Wiring), which learns network connectivity
through low-dimensional neuron embeddings (Figure~\ref{fig:method}).
Embedding distances determine which connections are active and modulate
their strengths, while network weights parameterize those connections. The embeddings and weights are optimized jointly from the prediction loss.

\subsection{Connectivity from neuron embeddings}

Each neuron $i$ in layer $\ell$ has a learned embedding
$\mathbf e_i^{(\ell)}\in\mathbb R^r$, where $r$ is the embedding dimension.
The output unit is treated as the final layer.
For a candidate connection from neuron $j$ in layer $\ell$ to neuron $i$ in
layer $\ell+1$, we define the connection score
\begin{equation}
\label{eq:lewi-score}
s_{ij}^{(\ell)}
=
R^2-
\left\|
\mathbf e_i^{(\ell+1)}-\mathbf e_j^{(\ell)}
\right\|_2^2.
\end{equation}
Thus, $R$ defines a connectivity radius: the score is positive when the two
embeddings are within distance $R$ and nonpositive otherwise.
The weights used to compute network activations in the forward pass, denoted
$\widetilde W_{ij}^{(\ell)}$, are obtained from learned base weights
$W_{ij}^{(\ell)}$ with a soft sigmoid gate
$g_{ij}^{(\ell)}=\sigma(s_{ij}^{(\ell)})$ and a binary mask
$m_{ij}^{(\ell)}=\mathbf 1\{s_{ij}^{(\ell)}>0\}$:
\begin{equation}
\label{eq:lewi-weight}
\widetilde W_{ij}^{(\ell)}
=
W_{ij}^{(\ell)}
g_{ij}^{(\ell)}
m_{ij}^{(\ell)}.
\end{equation}
The soft gate scales active weights, while the binary mask sets the
forward-pass weight to zero at or beyond radius $R$. 
We initialize embeddings i.i.d. as
$\mathbf e_i^{(\ell)}\sim\mathcal N(0,\sigma_e^2 I_r)$, where $\sigma_e$
is the standard deviation, and set
$R^2=\mathbb E_{\mathrm{init}}[\|\mathbf e_i-\mathbf e_j\|_2^2]
=2r\sigma_e^2$.
By default, $r=3$ and $\sigma_e=0.5$, giving $R^2=1.5$.
This places the connection threshold near the typical initial pairwise distance,
so the network starts with both active and inactive connections.
The radius remains fixed throughout training.

\subsection{Learning the wiring}

\begin{figure}[t]
  \centering
  \includegraphics[width=\linewidth]{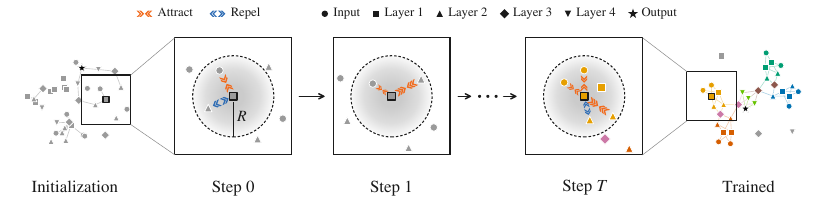}
  \vspace{-2em}
\caption{\textbf{LeWi learns connectivity through neuron embeddings.}
Schematic illustration of neuron positions and gradient-driven movement.
Adjacent-layer neurons connect within a fixed radius $R$ (dashed), with
shading indicating the soft gate. Prediction gradients can attract or repel
embeddings, changing connectivity as pairs cross the radius.
The zooms track one layer 1 neuron; trained colors illustrate post hoc
associations with target intermediates.
}
  \label{fig:method}
\end{figure}

The embeddings and weights are optimized jointly to minimize the
mean-squared prediction loss $\mathcal L$. During backpropagation,
we treat the binary masks as fixed and propagate gradients through the soft
gates of active connections.
Suppressing layer indices, let $\mathcal N(i)$ denote the neurons connected
to neuron $i$ by active connections, and let $W_{ij}$ denote the base weight of the corresponding active connection. The negative gradient with respect to
$\mathbf e_i$ is
\begin{equation}
-\nabla_{\mathbf e_i}\mathcal{L}
=
2\sum_{j\in\mathcal N(i)}
|W_{ij}|\sigma'(s_{ij})
\frac{\partial\mathcal{L}}
     {\partial|\widetilde W_{ij}|}
(\mathbf e_i-\mathbf e_j).
\end{equation}
For each $j\in\mathcal N(i)$, the same connection contributes an equal and
opposite term to the update of $\mathbf e_j$. If
$\partial\mathcal{L}/\partial|\widetilde W_{ij}|<0$, increasing the connection magnitude
reduces the loss, so gradient descent moves the embeddings closer. If
$\partial\mathcal{L}/\partial|\widetilde W_{ij}|>0$, decreasing the connection magnitude
reduces the loss, so the embeddings move apart. Each embedding update is the sum of attractive and repulsive
``forces'' from its active incoming and outgoing connections.
Figure~\ref{fig:method} illustrates this mechanism.

The binary mask converts this continuous movement into changes in
connectivity. When two embeddings reach or move beyond radius $R$, their
connection becomes inactive and receives no direct gradient. It can
reactivate if updates through other active connections later bring the
endpoints back within $R$. Masks are recomputed after each optimization
step.
Appendix~\ref{app:lewi-analysis} gives the full derivation.

\section{CSFRBench}
\label{sec:csfrbench}

CSFRBench is a controlled benchmark for evaluating whether gradient-based
learners exploit hidden compositional structure. It provides a target function
generator with known computation graphs and intermediate values, reference
networks with target-derived sparsity patterns, and a common evaluation
protocol. Learners receive only input--output examples, while the target computation graph and intermediate values
are retained for reference construction and diagnosis.
Together, these components enable joint evaluation of predictive sample
efficiency and learned computational structure.
Appendix~\ref{app:benchmark-design} summarizes the benchmark settings.

\begin{figure}
    \centering
    \includegraphics[width=\linewidth]{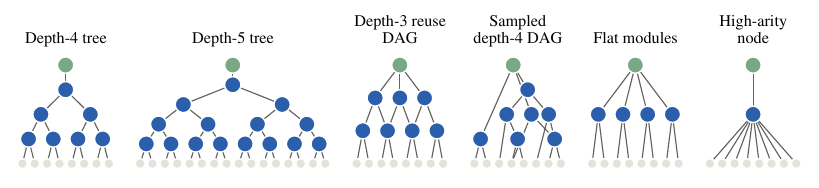}
    \vspace{-2.5em}
    \caption{
\textbf{Computation graphs in CSFRBench.}
Light nodes denote active inputs, blue nodes local functions, and green nodes
readouts. The figure shows the five fixed graph templates and one draw from
the sampled depth-4 DAG distribution.
    }
    \label{fig:graph-templates}
\end{figure}

\paragraph{Computation graphs.}
CSFRBench includes five fixed graph templates and a distribution of sampled
depth-4 DAGs (Figure~\ref{fig:graph-templates}). Together, these settings span
parallel low-arity computation, hierarchical composition, intermediate reuse,
varying graph topology, and a high-arity control.
In every graph, the readout is a standardized sum of its parent values rather
than a sampled local function.
Exact graph definitions are given in Appendix~\ref{app:graph-families}.

\paragraph{Local functions.}
Our primary local functions use an additive--multiplicative construction,
\begin{equation}
\label{eq:csfr-local-function}
g(z)
=
\sqrt{\frac{1-\rho}{k}}
\sum_{i=1}^{k} a_i(z_i)
+
\sqrt{\rho}
\prod_{i=1}^{k} b_i(z_i).
\end{equation}
By construction, the parent values $z_1,\ldots,z_k$ are independent because
they depend on disjoint input coordinates. Each $a_i$ and $b_i$ is sampled from
$\sin(\pi\nu u+2\pi\phi)$, with $\nu\sim U[0.25,0.40]$ and
$\phi\sim U[0,1)$, and standardized under the distribution of $z_i$.
Because $z_1,\ldots,z_k$ are independent, for
$k\geq2$ 
the additive and $k$-way interaction terms are therefore uncorrelated and
contribute variances $1-\rho$ and $\rho$, respectively.
We also use random Fourier functions \citep{rahimi2007random},
\begin{equation}
\label{eq:csfr-fourier}
g(z)=\frac{1}{\sqrt Q}\sum_{q=1}^{Q}
a_q\sin(\omega_q^\top z+\phi_q),
\end{equation}
where $Q$ is the number of terms, and $a_q$, $\omega_q$, and $\phi_q$
are sampled amplitudes, frequency vectors, and phases. Unlike the
additive--multiplicative construction, it does not prescribe
the relative contributions of additive effects and interactions.
Appendix~\ref{app:local-function-families} gives sampling distributions
and settings.

\paragraph{Target construction and characterization.}
Each target instance fixes its computation graph, local functions, and active
inputs across methods and sample sizes. Inputs have independent coordinates
$X_j\sim U[-1,1]$, and observations follow
$Y=f^\star(X)+\epsilon$, with $\epsilon\sim\mathcal N(0,\sigma^2)$.
Primary settings use only active inputs, while nuisance-input settings add
independent inert coordinates. Using a separate fixed normalization sample, we
center and scale each local output and the final readout to zero mean and unit
variance in topological order. The resulting normalization parameters are reused
across all training, validation, and test sets for that target instance.
On development instances, we characterize each target family's interaction
structure, input sensitivities, and output distribution, and test whether simpler
predictors can fit its input--output mapping. Final evaluation targets are sampled
after fixing the benchmark protocol, methods, hyperparameter spaces, and
statistical analysis. 
Appendices~\ref{app:target-generation}--\ref{app:local-function-capacity}
give details.

\paragraph{Graph-informed references.}
To assess the value of target connectivity, we use a family of reference
networks, GraphOracle($\ell$), with the same MLP backbone and fixed weight
masks derived from the first $\ell$ nonlinear levels of the target graph.
GraphOracle(1) uses the first level, while GraphOracle(all) uses the full
target connectivity. When nuisance inputs are present, GraphOracle(0)
provides a reference with known active-input support.
Appendix~\ref{app:reference-learners} details the mask construction.

\paragraph{Evaluation.}
Our primary outcome is predictive sample efficiency. At training-set size
$n$, we report normalized mean-squared error,
$\operatorname{NMSE}(n)=
\mathbb{E}_X[(\hat f_n(X)-f^\star(X))^2]/
\operatorname{Var}(f^\star(X))$.
We summarize learning curves by their area under the curve (AULC) on a
logarithmic training-set-size scale; lower values indicate better sample
efficiency (Appendix~\ref{app:predictive-evaluation}).

The known target computation also enables diagnostics of learned internal
organization. We measure linear decodability with ridge probes predicting each
target intermediate from each hidden layer and report held-out $R^2$
\citep{alain2016understanding}. Because linear decodability does not establish
that the decoded information is used in prediction \citep{belinkov2022probing},
we also assess counterfactual agreement under target-aligned interventions using
joint multi-variable Boundless DAS \citep{wu2023interpretability}. We report
held-out counterfactual $R^2$ between the learner's patched prediction and the
target's counterfactual output. Appendices~\ref{app:intermediate-probes}
and~\ref{app:causal-use} give the full protocols.

\section{Experiments}
\label{sec:experiments}

\subsection{Experimental setup}
\label{sec:experiment-setup}

\paragraph{Models and baselines.}
For CSFRBench, all methods use the same ReLU MLP backbone within each setting:
width 32 with four hidden layers, except for the depth-5 tree, which uses width
64 with eight hidden layers. We compare LeWi with an unregularized dense MLP
(Dense), $\ell_1$ and $\ell_2$ regularization, hard-concrete $L_0$
regularization \citep{louizos2017learning}, RigL \citep{evci2020rigging},
Fiedler regularization \citep{tam2020fiedler}, BIMT
\citep{liu2023seeing}, and learned neuron positions
\citep{erb2025training}. RigL adapts sparse connectivity during training,
Fiedler regularization and BIMT encourage structured wiring, and the
learned-neuron-position method parameterizes weights through neuron geometry.
CSFRBench architecture and training details are given in
Appendix~\ref{app:training-protocol}.

\paragraph{Training and model selection.}
Our primary settings use $d=8$ active inputs, local arity $k=2$, and
observation-noise variance $\sigma^2=0.01$. We train with Adam
\citep{kingma2014adam} using batch size 32 for 64,000 steps over
$n\in\{2^9,\ldots,2^{15}\}$. The depth-5 tree uses $d=16$,
$n\in\{2^{11},\ldots,2^{17}\}$, and 128,000 steps. Nuisance-input settings
use $d=32$, adding 24 or 16 inert inputs under the corresponding training
budget. For each target instance and training-set size, we evaluate 32
hyperparameter configurations from a scrambled Sobol sequence
\citep{sobol1967distribution,owen1995randomly}, using search spaces fixed on
development targets, and select the checkpoint and configuration with lowest
validation NMSE. Validation and test sets contain 4,096 and 65,536 examples
and are shared across methods. The selected model is evaluated once on the
test set. 
We measure sample efficiency by training-set size under this fixed
model-selection protocol, not total labeled-data requirements.
Details are given in Appendices~\ref{app:training-protocol}
and~\ref{app:hpo}.

\paragraph{Statistical reporting.}
We evaluate 16 target instances in each fixed-graph setting and 32
independently sampled graph/function instances in the sampled-graph setting.
We report mean AULC with two-sided 95\% Student-$t$ confidence intervals
over instances; method comparisons use paired per-instance AULC differences
with corresponding paired intervals. 
Comparisons whose intervals contain
zero are treated as inconclusive.
Appendix~\ref{app:statistical-reporting} gives the full procedure.

\subsection{Predictive sample efficiency}

\begin{figure}
\centering
\includegraphics[width=\linewidth]{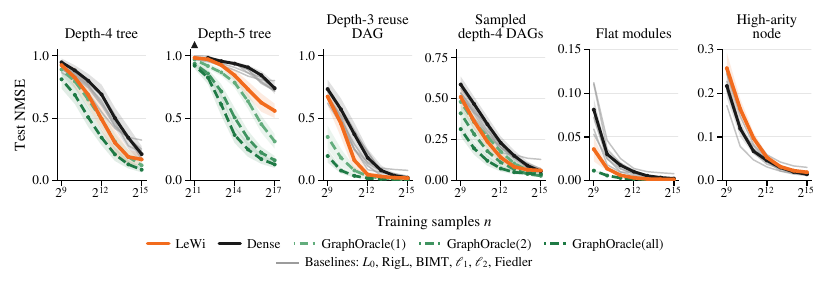}
\vspace{-2.5em}
\caption{
\textbf{LeWi improves sample efficiency across computational structures.}
Test NMSE versus training-set size for six representative CSFRBench settings.
Curves and bands show means and 95\% Student-$t$ confidence intervals over
target instances. Triangles mark means outside the displayed range. Baselines
are shown in gray. Complete results are in Appendix~\ref{app:additional-results}.
}
\label{fig:sample-efficiency}
\end{figure}

LeWi improves predictive sample efficiency across the five low-arity
CSFRBench settings in Figure~\ref{fig:sample-efficiency}, achieving the lowest
mean AULC among methods without access to target connectivity in each
setting. The same ordering holds in two Fourier settings omitted from
the figure. Appendix~\ref{app:complete-predictive} reports full learning
curves and paired numerical comparisons for all settings. Relative to
Dense, LeWi lowers mean AULC by 20\% on the depth-4 tree, 34\% on the
reuse DAG, and 56\% on flat modules. Across 32 independently sampled
depth-4 DAGs, mean AULC is 27\% lower. On the larger depth-5 tree,
LeWi achieves 32\% lower mean AULC than Dense, while Dense performs
better on the high-arity node. These results are consistent with LeWi benefiting from compositional sparsity across different computational structures.

Graph-informed references quantify the additional sample-efficiency
benefit available from target connectivity. On the depth-4 tree,
LeWi approaches GraphOracle(1) and closes $51\%$ of the AULC gap
between Dense and GraphOracle(all), compared with $26\%$ for $L_0$.
LeWi remains behind GraphOracle(1) on the reuse DAG and depth-5 tree,
despite outperforming all learned baselines in both settings.
Thus, learned wiring captures part of the predictive benefit of target
connectivity without access to the target graph, while the remaining
oracle gaps indicate room for improvement, particularly on deeper
and reuse-based compositions.

\subsection{Learned computational structure}
\label{sec:learned-structure}

\begin{figure}
    \centering
    \includegraphics[width=\linewidth]{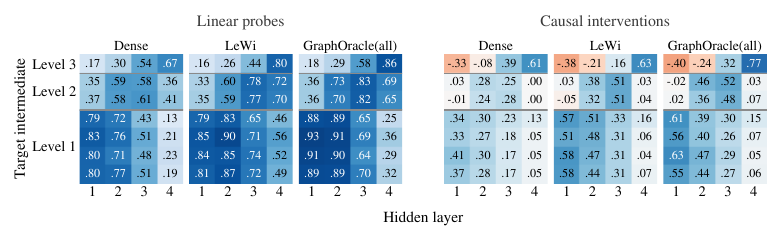}
    \vspace{-2.5em}
    \caption{
        \textbf{LeWi develops stronger target-aligned representations.}
        Left: held-out $R^2$ from linear probes predicting each target intermediate
        from each hidden layer. Right: held-out counterfactual $R^2$ under joint
        multi-variable Boundless DAS. Rows are grouped by target level. Results are for the depth-4 tree at
        $n=2^{14}$, averaged over 16 target instances.
    }
    \label{fig:learned-structure}
\end{figure}

We next examine whether LeWi's predictive gains are accompanied by
hidden representations aligned with the target computation
(Figure~\ref{fig:learned-structure}). On the depth-4 tree, linear
decodability follows the target hierarchy: lower-level intermediates are
most decodable in earlier hidden layers, while deeper intermediates are
most decodable in later hidden layers. LeWi strengthens this organization
relative to Dense. Using the maximum mean probe $R^2$ across hidden layers,
LeWi closes more than half of the Dense-to-GraphOracle(all) gap for level-1
intermediates, roughly three quarters for level-2 intermediates, and about
two thirds for the level-3 intermediate.

Linear decodability alone does not establish whether these
representations are used in predictions. We therefore use joint
multi-variable Boundless DAS to test whether interventions on
target-aligned hidden subspaces produce the corresponding
counterfactual predictions. The intervention results show the same
hierarchical organization. For level-1 intermediates in the first
hidden layer, LeWi closes nearly all of the Dense-to-GraphOracle(all)
gap, while level-2 intermediates in the third layer reach approximately
GraphOracle(all) performance. Intervention gains are largest at lower
and intermediate target levels, while LeWi and Dense are similar for
the level-3 intermediate. 
Together, the diagnostics show that LeWi's
hidden representations are more closely aligned with the target
computation than Dense.
Appendix~\ref{app:representation-diagnostics} extends these diagnostics
to additional learned baselines, training-set sizes, and target families.

\subsection{Input selection, wiring alignment, and ablations}
\label{sec:wiring-analysis}

\begin{figure}[t]
    \centering
    \includegraphics[width=\linewidth]{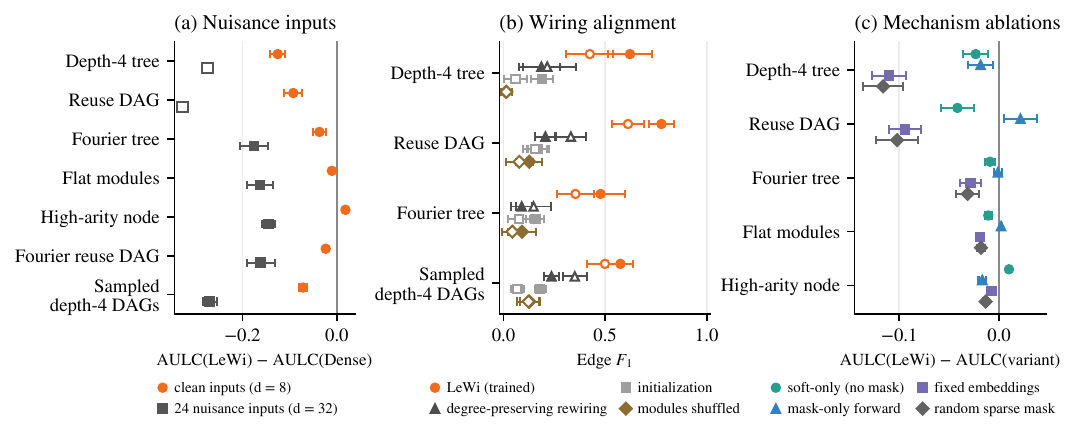}
    \vspace{-2em}
\caption{
\textbf{Input selection, wiring alignment, and mechanism ablations.}
(a) LeWi--Dense paired AULC differences with all inputs relevant or 24
additional inert inputs. Hollow squares show medians when divergent Dense runs
place means outside the plotted range.
(b) Edge F1 over all target edges (filled) and edges from target intermediates
(hollow) for trained LeWi and controls at $n=2^{15}$.
(c) LeWi--variant paired AULC differences for four mechanism ablations.
Negative differences favor LeWi in (a) and (c). Error bars show 95\%
Student-$t$ confidence intervals over target instances.
}
\label{fig:wiring-analysis}
\end{figure}

\paragraph{Input selection.}
LeWi provides an additional benefit when irrelevant inputs are present:
its sample-efficiency advantage over Dense increases across all seven
settings after adding 24 inert inputs
(Figure~\ref{fig:wiring-analysis}a). The effect is especially notable
on the high-arity node, where Dense performs better when all inputs are
relevant but falls behind with nuisance inputs. A plausible
explanation is that learned gates and masks suppress connections from
irrelevant inputs, shifting part of the input-selection burden from
weight optimization to connectivity learning. 
Appendix~\ref{app:nuisance-input-results} reports the complete
nuisance-input learning curves and paired comparisons.

\paragraph{Wiring alignment.}
LeWi's learned connectivity also aligns with dependencies in the target
computation graph. We extract a module graph without target information by
clustering neurons on active input--output paths according to embedding
distance, using average linkage with a cut at radius $R$. We connect two
modules when more than half of their candidate connections
are active. Modules are then matched post hoc to target intermediates using
ridge probes, and the resulting dependency graph is compared with the target
graph using edge precision, recall, and $F_1$. Because target intermediates
are used for post hoc matching, these metrics measure structural agreement
with the target graph rather than graph recovery.
Appendix~\ref{app:structure-discovery} gives the full procedure,
representative extracted graphs, and additional checks of the
module-extraction procedure.

Across four structured settings, trained LeWi achieves more than three times
the edge $F_1$ of initialization and randomized controls on average
(Figure~\ref{fig:wiring-analysis}b). Alignment is strongest on the reuse DAG
but remains well above the controls in every setting. For edges originating
at target intermediates, including edges to the output, mean $F_1$ remains
$0.473$, showing alignment beyond input selection. Precision exceeds recall
in every setting, so missed target edges are more common than spurious edges.
Some missed edges arise when a target intermediate has no matched module.
Because connectivity is parameterized by low-dimensional neuron embeddings,
LeWi also provides a direct visualization of its learned wiring.
Figure~\ref{fig:embedding-geometry} shows examples across target structures.

\paragraph{Mechanism ablations.}
We examine which components of LeWi contribute to its sample-efficiency gain.
Freezing the embeddings or replacing learned connectivity with a fixed random
sparse mask degrades performance on compositional targets
(Figure~\ref{fig:wiring-analysis}c), showing that sparsity alone is
insufficient. Removing the hard mask also hurts
these targets but helps on the high-arity node. In contrast, a mask-only variant,
which uses the learned binary mask without soft gate scaling in the forward
pass while retaining embedding learning, performs similarly to LeWi.
Together, these ablations indicate that adapting which connections are active
matters more than continuously scaling active connections.
LeWi also reaches lower connection density than at initialization on
compositional targets, especially with nuisance inputs, while the high-arity
control remains dense. Appendices~\ref{app:connection-density}--\ref{app:computational-cost}
report connection density, parameter sensitivity, and computational cost.

\subsection{External validation on world-model prediction}
\label{sec:external-validation}

\begin{figure*}[t]
    \centering
    \includegraphics[width=0.85\textwidth]{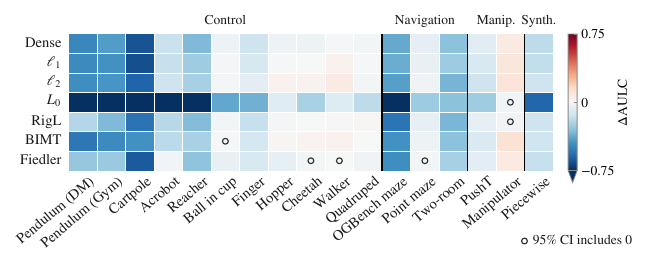}
    \vspace{-1.5em}
\caption{\textbf{World-model prediction.}
Paired $\Delta\mathrm{AULC}
=\mathrm{AULC}(\mathrm{LeWi})-\mathrm{AULC}(\mathrm{baseline})$
across 17 tasks and seven baselines. Negative values favor LeWi.
LeWi wins 99 of 119 comparisons, ties 6, and loses 14.
Hollow circles indicate 95\% confidence intervals containing zero.
Colors are clipped at $\pm0.75$.
Tasks are grouped by environment type and ordered by state dimension.
Appendix~\ref{app:world-model-results} reports complete results, including
the learned-neuron-position comparison.
}
    \label{fig:world-models}
\end{figure*}

To test whether this inductive bias transfers beyond controlled targets,
we evaluate LeWi on state-based prediction tasks from the
\texttt{stable-worldmodel} suite \citep{maes2026stable}. We evaluate the same
set of methods as in CSFRBench using a width-256, four-hidden-layer ReLU MLP.
To focus on tasks learnable with this architecture, we retain 17 of the 21
state-based, continuous-action environments that satisfy fixed,
LeWi-independent criteria for nontrivial prediction, a data-limited regime,
and reliable simulator replay. The selected tasks span control, navigation,
articulated and contact dynamics, manipulation, and a synthetic piecewise
system. Models predict state changes and are evaluated by 16-step open-loop
rollout over seven training-set sizes.
Appendices~\ref{app:world-model-tasks}--\ref{app:world-model-results} give the
task selection, experimental protocol, and complete results.

Across the 119 task--baseline pairs (17 tasks $\times$ 7 baselines), LeWi wins
99, ties 6, and loses 14 (Figure~\ref{fig:world-models}); against Dense, it
wins 16 of 17 tasks. Gains are strongest and most consistent on low-dimensional control and
navigation tasks, while most ties and losses occur on articulated-control tasks
with many coupled joints and state variables, and on manipulation, which also
couples robot and object dynamics. Together with the CSFRBench results, this
pattern is consistent with LeWi benefiting most when predictive dependencies
can be organized into sparse, low-arity interactions, with less advantage as
those dependencies become more broadly coupled.
\section{Discussion}
\label{sec:discussion}

Our results support treating connectivity, alongside weights, as a learned
model component for sample-efficient prediction. CSFRBench enables joint
evaluation of predictive sample efficiency and learned computational
structure. 
LeWi improves sample efficiency on targets with hierarchical
structure, intermediate reuse, or independent modules, and
improves prediction on state-based world-model tasks. It also
develops representations and connectivity aligned with the target
computation, while its low-dimensional neuron embeddings facilitate
visualization and interpretation of the learned computational structure.
Inspired by the spatial organization of biological nervous systems,
LeWi jointly learns neuron embeddings that organize connectivity and
connection weights. This draws an analogy to coupling the evolution
and development of neural circuitry with experience-dependent
synaptic learning
\citep{wang2016influence,cang2013developmental,holtmaat2009experience}.

This approach also connects to structure and parameter learning in
probabilistic graphical models \citep{koller2009probabilistic}.
In Bayesian networks, graph structure specifies the factorization
of a joint distribution, while parameters specify the local
conditional distributions. LeWi similarly learns connectivity and
local computations jointly, but its graph describes computational
dependencies among learned representations rather than probabilistic
dependencies among random variables. This suggests exploring
geometric connectivity as an alternative parameterization for
graphical-model structure learning, with appropriate probabilistic
objectives and structural constraints.

Several limitations remain. The remaining gap to GraphOracle indicates room to better
exploit target connectivity. LeWi's random initialization and hard
masks can make initially inactive connections difficult to recover.
Its fixed-radius, adjacent-layer connectivity restricts the range
of learnable wiring patterns. LeWi also evaluates all candidate
connections densely, so learned sparsity does not reduce training
computation. With optimized CPU implementations, its runtime is
1.32 and 1.02 times that of Dense on the $32\times4$ and
$64\times8$ backbones, respectively
(Appendix~\ref{app:computational-cost}).

Future work could improve initialization and rewiring to recover
better computational structures, and exploit learned sparsity to
reduce training cost. Inspired by the distinct spatial organization of axons and dendrites
in biological neurons, learning separate embeddings for outgoing
and incoming connections could enable sparse long-range connectivity
and support learning larger computational graphs. Further biological
mechanisms may offer additional inspiration for learning connectivity.
Extending LeWi to Transformers would test whether geometric
connectivity benefits attention-based architectures, while input-
or task-dependent connectivity could enable conditional computation
and module reuse in continual learning. Finally, extending CSFRBench
to larger graphs, richer local functions, and sequential settings
would help assess scalability and adaptation.

\subsection*{AI use statement}

We used generative AI tools to provide feedback on research methodology
and experimental design, assist with interpretation of experimental
results, and help develop and check mathematical derivations and claims.
We also used generative AI for literature search and summarization,
manuscript drafting and editing, and LaTeX assistance. An AI coding agent
was used to write and modify experiment, synthetic-data-generation,
analysis, and verification code. All AI-assisted work was reviewed by the
authors: mathematical claims were checked manually, cited literature was
verified against the original sources, and generated or modified code was
tested against experiment records and independent verification checks.
We take responsibility for the final content of this work, including text,
claims, code, and artifacts produced with the aid of generative AI.

\subsection*{Ethics statement}

This work studies general-purpose learning methods using synthetic regression
targets and simulated world-model prediction environments. It involves no
human subjects, personal or sensitive data, or deployment in high-stakes
settings.

\subsection*{Reproducibility statement}

Appendix~\ref{app:lewi-analysis} specifies the LeWi implementation and
analysis. Appendix~\ref{app:benchmark-details} gives the complete CSFRBench
construction, training, hyperparameter-search, evaluation, and statistical
protocols. Appendix~\ref{app:additional-results} reports additional results
and controls, and Appendix~\ref{app:world-models} documents the world-model
task suite, data collection, model selection, search spaces, and complete
results. Source code for LeWi and CSFRBench, experiment configurations, and scripts for reproducing
the reported analyses are available on the project page:
\url{https://zhou-hangyu.github.io/lewi/}.

\subsubsection*{Acknowledgments}
We thank Jixian Su, Anisha Pal, and Kangping Hu for helpful discussions and feedback. This research was supported in part by computing resources provided by the Partnership for an Advanced Computing Environment (PACE) at the Georgia Institute of Technology (RRID:SCR\_027619).

\bibliography{iclr2027_conference}
\bibliographystyle{iclr2027_conference}

\clearpage
\appendix
\section{Statistical background}
\label{app:statistical-background}

The statistical motivation for CSFR comes from nonparametric regression rates
for compositional functions. Consider a target
$f^\star=g_q\circ\cdots\circ g_0$ whose level-$i$ component functions are
$\beta_i$-Hölder smooth and depend on at most $t_i$ variables, with
$t_i\ll d$.
\citet{schmidt2020nonparametric} define the effective smoothness
$\beta_i^\star=\beta_i\prod_{\ell=i+1}^q(\beta_\ell\wedge 1)$ and show that,
under their assumptions, the minimax prediction rate is governed up to
logarithmic factors by
\[
\phi_n
=
\max_{i=0,\ldots,q}
n^{-\frac{2\beta_i^\star}{2\beta_i^\star+t_i}}.
\]
A sparse ReLU estimator based on near-global empirical-risk minimization attains
$O(\phi_n\log^3 n)$, matching the minimax exponent. In contrast, generic
$d$-dimensional $\beta$-Hölder regression has the classical rate
$n^{-2\beta/(2\beta+d)}$ \citep{stone1982optimal}.
This rate shows that statistical complexity can depend on local arity and
smoothness rather than ambient dimension. However, it does not establish that
standard gradient-based training discovers and exploits unknown compositional
structure.

\section{LeWi implementation and analysis}
\label{app:lewi-analysis}

\subsection{Implementation details}
\label{app:lewi-implementation}
Unless otherwise specified, LeWi uses embedding dimension $r=3$ and
$\sigma_e=0.5$, giving $R^2=2r\sigma_e^2=1.5$. The same fixed radius is
used for every affine layer, including the readout. Binary masks are
recomputed after every optimizer step and treated as fixed during
backpropagation.

We use two rejection-sampling checks at initialization. If a layer
transition contains no active connection, we redraw its receiver
embeddings until at least one connection is active. We also redraw any
input embedding with fewer than five active outgoing connections until
it has at least five. These checks apply only at initialization; no
degree constraint is imposed during training. LeWi otherwise uses no
target sparsity level or connectivity penalty.
These checks use no target information and apply identically to all input coordinates.

\subsection{Properties of the parameterization}

\paragraph{Shared score structure.}
For one layer, let $X\in\mathbb R^{m\times r}$ and
$Y\in\mathbb R^{n\times r}$ contain the embeddings of the target and source
neurons. Expanding the squared distance gives
\[
S
=
2XY^\top
+
(R^2\mathbf 1-a)\mathbf 1^\top
-
\mathbf 1b^\top,
\]
where $a_i=\|x_i\|_2^2$ and $b_j=\|y_j\|_2^2$. Hence
$\operatorname{rank}(S)\le r+2$, and after removing row and column means the
rank is at most $r$. Thus, LeWi generates all connection scores in a layer from
shared $r$-dimensional neuron embeddings rather than learning one score per
connection independently.
The low-rank structure applies only to the score matrix: the connection weights \(W\) are unconstrained, and the thresholded mask need not be low rank.

\paragraph{Binary mask.}
Without the binary mask, making the soft gate smaller than
$\epsilon<1/2$ requires
\[
d^2
\ge
R^2+\log\frac{1-\epsilon}{\epsilon},
\]
so exact disconnection is reached only as the distance diverges. With the
binary mask, a connection becomes exactly zero once its embedding distance
reaches $R$.

\section{CSFRBench construction and evaluation protocol}
\label{app:benchmark-details}

\subsection{Benchmark design and evaluation settings}
\label{app:benchmark-design}

CSFRBench is a controlled benchmark for studying whether gradient-based
learners exploit hidden compositional structure. Each target has a known
computation graph, local functions, and intermediate values, but learners
receive only input--output examples. This makes it possible to evaluate
predictive sample efficiency together with the internal representations,
counterfactual agreement, and connectivity that emerge during training.

The benchmark varies both computation graph and local-function family.
Unless otherwise specified, the input dimension is $d=8$, and the observation-noise variance is $\sigma^2=0.01$; compositional
targets use local arity two. The depth-5 tree uses $d=16$. Target outputs
are standardized to mean zero and variance one, making the noise level
comparable across target families. Table~\ref{tab:csfrbench-settings}
summarizes the evaluation settings.

\begin{table}[t]
\centering
\caption{\textbf{CSFRBench evaluation settings.}
The suite spans parallel, hierarchical, reuse-based, sampled, and high-arity
computation, together with an independent local-function family and a
depth/scale extension.}
\label{tab:csfrbench-settings}
\footnotesize
\setlength{\tabcolsep}{4pt}
\begin{tabular}{@{}lllcl@{}}
\toprule
Setting & Local functions & Structure & Instances & Purpose \\
\midrule
Single local function
& Add.--mult. & arity 2 & 16
& local fitting and input selection \\

Flat modules
& Add.--mult. & four disjoint pairs & 16
& parallel low-arity computation \\

Depth-4 tree
& Add.--mult. & balanced tree & 16
& hierarchical composition \\

Reuse DAG
& Add.--mult. & reused intermediates & 16
& composition with intermediate reuse \\

Sampled depth-4 DAGs
& Add.--mult. & sampled topology & 32
& generalization across graph draws \\

Depth-4 tree
& Fourier & balanced tree & 16
& independent local-function family \\

Reuse DAG
& Fourier & reused intermediates & 16
& independent local-function family \\

High-arity node
& Fourier & arity 8 & 16
& high-arity control \\

High-arity node
& Add.--mult. & arity 8 & 16
& local-function hardness stress setting \\

Depth-5 tree
& Add.--mult. & balanced tree & 16
& depth/scale extension \\
\bottomrule
\end{tabular}
\end{table}

These settings span distinct structural regimes rather than forming a matched
factorial design.

\subsection{Computation graphs}
\label{app:graph-families}

CSFRBench includes fixed graph templates and a distribution of sampled graphs.
The fixed templates in Figure~\ref{fig:graph-templates} span parallel
low-arity computation, hierarchical composition, intermediate reuse, and a
high-arity control. In every graph, the readout is a standardized sum of its
parent values, rather than a sampled local function.

\paragraph{Flat modules.}
Four disjoint two-input local functions operate on eight active inputs,
followed by a standardized additive readout. This setting contains parallel
low-arity computations without hierarchical composition.

\paragraph{Depth-4 tree.}
The primary hierarchical setting is a balanced binary tree with eight active
inputs and three nonlinear levels containing $4$, $2$, and $1$ local
functions. Every local function has arity two, and the readout standardizes the
root value.

\paragraph{Reuse DAG.}
The fixed DAG uses eight active inputs and seven two-input local functions. Its
first level computes $z_1=g_1(x_{i_1},x_{i_2})$,
$z_2=g_2(x_{i_3},x_{i_4})$, $z_3=g_3(x_{i_5},x_{i_6})$, and
$z_4=g_4(x_{i_7},x_{i_8})$. Its second level computes
$z_5=g_5(z_1,z_2)$, $z_6=g_6(z_2,z_3)$, and
$z_7=g_7(z_3,z_4)$, followed by a standardized additive readout of
$z_5,z_6,z_7$. Thus, $z_2$ and $z_3$ are each reused by two downstream
local functions. The DAG matches the depth-4 tree in active-input count,
local-function count, and local arity, but differs in depth, connectivity,
reuse, and readout; their comparison therefore does not isolate reuse.

\paragraph{High-arity node.}
The high-arity setting applies a single eight-input local function directly to
all active variables, removing the low-arity compositional decomposition.

\begin{figure}[t]
\centering
\includegraphics[width=\linewidth]{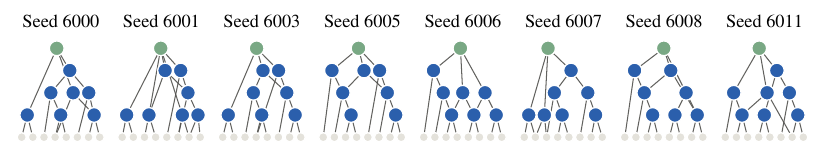}
\caption{
\textbf{Sampled computation graphs.}
Eight distinct topologies drawn from the depth-4 graph distribution. Each graph
has eight active inputs, seven arity-two local functions, and three nonlinear
levels, while branching, intermediate reuse, and readout connectivity vary.
}
\label{fig:held-out-graphs}
\end{figure}

\paragraph{Sampled depth-4 DAGs.}
Each instance independently samples a connected graph with eight active inputs,
seven arity-two local functions, and three nonlinear levels followed by a
standardized additive readout. The sampler constructs two or three input
strands and adds reuse connections while requiring that the parents of each
local function have disjoint input ancestry and that no unordered parent pair
is duplicated. Graph structure is sampled independently of the local-function
parameters. Across 4,096 development draws, the sampler produces 354 distinct canonical
topologies; a typical 32-instance cohort contains about 30 distinct topologies.
Figure~\ref{fig:held-out-graphs} shows eight distinct sampled topologies.

\paragraph{Depth-5 tree.}
The depth/scale extension is a balanced tree with 16 active inputs and four
nonlinear levels containing $8$, $4$, $2$, and $1$ local functions.

\subsection{Local-function families}
\label{app:local-function-families}

CSFRBench uses two smooth local-function families with different interaction
structure.

\paragraph{Additive--multiplicative functions.}
For parents $z_1,\ldots,z_k$, we define
\[
g(z_1,\ldots,z_k)
=
\sqrt{\frac{1-\rho}{k}}
\sum_{i=1}^k a_i(z_i)
+
\sqrt{\rho}
\prod_{i=1}^k b_i(z_i),
\]
where each one-dimensional factor is centered and standardized under the
induced distribution of its parent. With independent standardized parents,
the additive term contributes first-order variation while the product
introduces a $k$-way interaction. We use $\rho=0.5$ in the primary
experiments.

The factors are sinusoidal. Before standardization,
\[
\widetilde s(u)
=
\sin\!\left(2\pi\left(fu/2+\phi\right)\right),
\qquad
f\sim\mathrm U[0.25,0.4],\quad
\phi\sim\mathrm U[0,1),
\]
and
\[
s(u)
=
\frac{\widetilde s(u)-\mathbb E[\widetilde s(U)]}
{\sqrt{\operatorname{Var}[\widetilde s(U)]}}.
\]
At leaf nodes we use $u=\sqrt{3}x$ for $x\sim\mathrm U[-1,1]$; at deeper
nodes, the parent values are the sequentially normalized intermediates
described above.

\paragraph{Random Fourier functions.}
As a complementary local-function family, we use
\[
g_v(z)
=
\frac{1}{\sqrt Q}
\sum_{q=1}^Q
a_{v,q}\sin(\omega_{v,q}^{\top}z+\phi_{v,q}),
\]
with
$a_{v,q}\sim\mathrm U[0.7,1]$,
$\phi_{v,q}\sim\mathrm U[0,2\pi)$, and
$\omega_{v,q}\sim\mathcal N(0,\sigma_\omega^2 I_k/k)$.
For arity-two functions we use $Q=8$ and $\sigma_\omega=0.85$.

For the arity-eight Fourier function, we use $Q=24$ and
$\sigma_\omega=1.5$. This setting remains learnable in isolation and therefore
serves as the high-arity control. The arity-eight additive--multiplicative
setting is substantially harder to fit locally and is retained as a stress
setting rather than a decomposition control.

\subsection{Target generation and sequential normalization}
\label{app:target-generation}

Each target instance fixes its computation graph, local-function parameters,
active input coordinates, and normalization transforms. These quantities are
sampled once and then held fixed across methods, training-set sizes, and
train/validation/test datasets. Separate deterministic random streams govern
the target structure and local functions, the assignment of active variables
to ambient coordinates, and the normalization pool.

Inputs have independent coordinates $X_j\sim\mathrm{U}[-1,1]$. For each target
instance, we propagate a deterministic pool of $2^{18}$ inputs through the
computation graph in topological order. At each local function, its output is
centered and standardized under the induced parent distribution, and the stored
affine transform is then used for all downstream computation and all later
datasets. The final readout is standardized in the same way. Thus, every target
intermediate and the final target have approximately zero mean and unit variance
under the instance-specific normalization distribution.

The stored normalization transforms are part of the target instance and are
never recomputed per method, training-set size, or dataset. A local-function
draw is redrawn only if its raw standard deviation on the normalization pool is
below $10^{-3}$ or its output is non-finite; otherwise no target-level
characterization statistic is used to accept or reject individual instances.

Training observations follow
$Y=f^\star(X)+\epsilon$, with
$\epsilon\sim\mathcal N(0,\sigma^2)$. Unless otherwise specified,
$\sigma^2=0.01$.

\subsection{Target characterization}
\label{app:target-characterization}

Before final evaluation, we characterize each target family on independently
sampled development instances. Diagnostics include parent-wise sensitivity,
functional-ANOVA and low-order Sobol indices where applicable
\citep{sobol1990sensitivity,homma1996importance}, input-gradient statistics,
and the distributions and tails of target intermediates and outputs. We also
fit linear and pairwise-additive predictors using the true active inputs to
detect target families whose input--output mapping is substantially simpler
than the intended composition.

These diagnostics are used to characterize and, when necessary, revise a
target family before final evaluation; they are not used to select individual
final instances. Per-instance checks are limited to generation correctness,
including finite values, required graph constraints, and numerical
nondegeneracy.

Target-family choices are fixed using development targets without LeWi
results. Final evaluation targets are sampled only after the benchmark
protocol, comparison set, hyperparameter search spaces, and statistical
analysis are fixed.

\subsection{Local-function fitting}
\label{app:local-function-capacity}

We fit each local-function family in isolation to separate local approximation
from input selection and composition. Models receive exactly the arguments of
the local function ($d=k$), with no inert inputs or observation noise. We
evaluate 16 development draws per family.

\begin{figure}[t]
\centering
\includegraphics[width=0.9\linewidth]{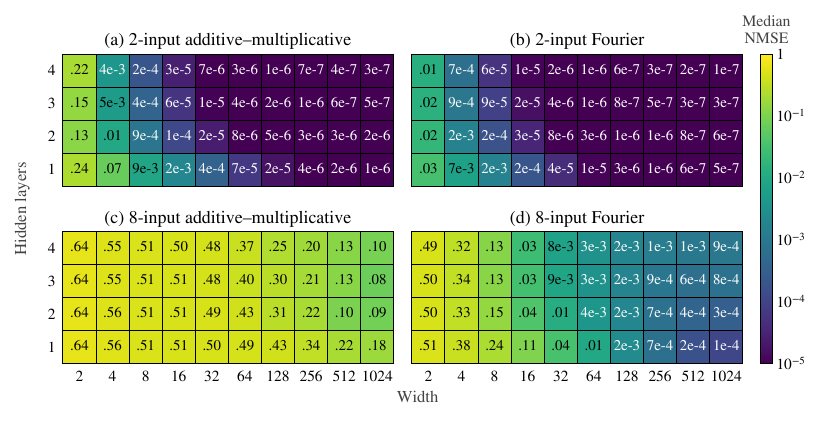}
\caption{
\textbf{Local-function fitting capacity.}
Median held-out NMSE over 16 development draws as a function of MLP width and
hidden depth. Each model receives exactly the local-function arguments.
(a,b) Arity-two local-function families.
(c) Arity-eight additive--multiplicative stress setting.
(d) Arity-eight Fourier control.
}
\label{fig:local_function_capacity}
\end{figure}

We fit ReLU MLPs with one to four hidden layers and widths
$W\in\{2,4,8,16,32,64,128,256,512,1024\}$. Each model is trained with Adam
for 32,000 steps on 65,536 training examples, using eight scrambled-Sobol
learning rates from $[10^{-4},10^{-1}]$; model selection uses 16,384 validation
examples and evaluation uses 65,536 test examples.

As shown in Figure~\ref{fig:local_function_capacity}, the arity-two functions
are accurately fit at modest capacity: a one-hidden-layer width-8 MLP achieves
median NMSE $0.0085$ for the additive--multiplicative family and $0.0016$ for
the Fourier family. The arity-eight Fourier function is also accurately
learnable at moderate capacity, whereas the arity-eight
additive--multiplicative function remains substantially harder even at much
larger widths. We therefore use the former as the high-arity control and retain
the latter only as a stress setting.

\subsection{Graph-informed references}
\label{app:reference-learners}

GraphOracle($\ell$) uses the same MLP backbone and training protocol as the
learned methods, but fixes its connectivity using the first $\ell$ nonlinear
levels of the target computation graph. GraphOracle(0) supplies only the active
input variables, while GraphOracle(all) supplies the full target connectivity.
Intermediate variants progressively supply additional nonlinear levels.

\begin{figure}[t]
\centering
\includegraphics[width=0.75\linewidth]
{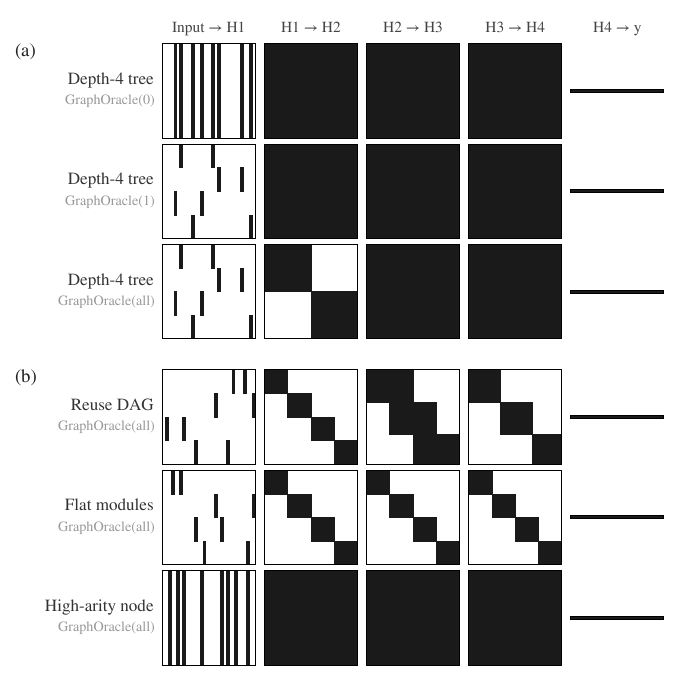}
\caption{
\textbf{GraphOracle connectivity masks.}
Black entries denote retained weights.
(a) GraphOracle(0), GraphOracle(1), and GraphOracle(all) on the depth-4 tree.
(b) GraphOracle(all) on the reuse DAG, flat modules, and high-arity node.
}
\label{fig:graph-oracle-masks}
\end{figure}

To construct the masks, hidden units are partitioned into blocks associated
with target intermediates. A block receives connections only from blocks
corresponding to its supplied target parents; values reused downstream are
carried forward and may fan out to multiple consumers. For a target with $L$
nonlinear levels and an MLP with $H\ge L$ hidden layers, level $r$ is assigned
\[
s_r
=
\left\lfloor\frac{rH}{L}\right\rfloor
-
\left\lfloor\frac{(r-1)H}{L}\right\rfloor
\]
consecutive hidden layers.
Values that must pass through an additional ReLU layer receive at least two
units so that a signed scalar can be represented. Connections beyond the
supplied prefix are left unrestricted.
Masked layers are initialized according to their retained fan-in; otherwise
the architecture, optimizer, and training protocol are unchanged.
Figure~\ref{fig:graph-oracle-masks} shows representative masks.
GraphOracle is a diagnostic reference rather than a matched causal intervention, since masking changes both the hypothesis class and initialization.

\subsection{Predictive evaluation}
\label{app:predictive-evaluation}

For each target instance and training-set size, methods use the same
training, validation, and test data. Validation and test sets contain
4,096 and 65,536 examples, respectively, and are fixed across
training-set sizes. After model selection
(Appendix~\ref{app:hpo}), the selected model is evaluated once on the
test set against the noise-free target $f^\star(X)$.

At training-set size $n$, we report normalized mean-squared error,
\[
\operatorname{NMSE}(n)
=
\frac{\mathbb E[(\hat f_n(X)-f^\star(X))^2]}
{\operatorname{Var}(f^\star(X))}.
\]
For sample sizes $n_1<\cdots<n_K$, we summarize the learning curve by
\[
\operatorname{AULC}
=
\frac{1}{\log_2 n_K-\log_2 n_1}
\sum_{j=1}^{K-1}
\frac{\operatorname{NMSE}(n_j)+\operatorname{NMSE}(n_{j+1})}{2}
\left(\log_2 n_{j+1}-\log_2 n_j\right).
\]
Lower AULC indicates better sample efficiency. Because the validation
set is fixed across $n$, these curves measure performance as a function
of training-set size under a fixed model-selection protocol, rather
than total labeled-data requirements.

\subsection{Linear-probe protocol}
\label{app:intermediate-probes}

Each target instance has an $8{,}192$-example diagnostic pool, split
deterministically into $4{,}096$ probe-training, $2{,}048$ probe-validation,
and $2{,}048$ probe-test examples. For each target intermediate $z_v$ and
learner hidden layer $h^{(\ell)}$, we fit a ridge-regression probe to predict
$z_v$ from $h^{(\ell)}$.

Learner features are standardized and probe targets are centered using
probe-training statistics only. Hidden features active on fewer than 10
probe-training examples are excluded before standardization. We select
$\lambda\in\{10^{-4},10^{-3.5},\ldots,10^7\}$ by validation error and
evaluate the selected probe once on the probe-test split without refitting.
We report ordinary, unclipped held-out $R^2_{v,\ell}$ as a measure of
linear decodability of target intermediate $v$ from hidden layer $\ell$.

\subsection{Counterfactual-intervention protocol}
\label{app:causal-use}

We use joint multi-variable Boundless Distributed Alignment Search (DAS)
\citep{wu2023interpretability} to test whether target intermediates can be
aligned with interventionally effective subspaces of a learner's hidden
state. For each learner and hidden layer, DAS learns an orthogonal rotation
whose coordinates are partitioned into disjoint blocks associated with the
target intermediates.

For a base/source pair and target intermediate $v$, we replace only the
block assigned to $v$ in the rotated base representation with the
corresponding source block, invert the rotation, and continue the learner
forward pass. We compare the resulting final prediction with the target's
counterfactual final output obtained by replacing $v$ with its source value
and recomputing its descendants in the target graph. Learner parameters
remain fixed throughout.

We report ordinary, unclipped held-out counterfactual $R^2$. We probe hidden
layers 1--4 using 10,000/1,000/2,000 counterfactual pairs per target
intermediate for training/validation/test. DAS is optimized for 18 epochs
with Adam, batch size 64, rotation learning rate $0.003$, boundary learning
rate $0.0003$, and validation every 200 updates. We use three restarts and
select the restart and learner layer on validation data before one test
evaluation.

The implementation uses half-offset boundary coordinates, a zero-capable
width clamp, shared-budget simplex rescaling, and temperature scaling by
$d/4096$. Protocol constants were fixed on development targets using planted
networks and GraphOracle controls. The protocol is calibrated for the
width-32, four-hidden-layer backbone and is therefore not applied to the
depth-5 setting.

This diagnostic tests counterfactual agreement under a jointly fitted,
target-aligned subspace decomposition. Failure to find such a decomposition
does not rule out other, potentially overlapping or nonlinear,
representations of the target intermediates.

\subsection{Learner architecture and training}
\label{app:training-protocol}

The local-function fits in Figure~\ref{fig:local_function_capacity} show that
the arity-two constituent functions are accurately approximated at modest
capacity. For all settings except the depth-5 tree, we use a ReLU MLP with
width 32 and four hidden layers. The depth-5 tree uses width 64 and eight
hidden layers. All methods use the same backbone within each setting.

Models are trained with Adam using $\beta_1=0.9$, $\beta_2=0.999$,
$\epsilon=10^{-8}$, batch size 32, and zero optimizer weight decay.
The initial learning rate is selected by hyperparameter search and
follows cosine decay to 1\% of its initial value. We use no warmup or
early stopping. The eight-input settings use 64,000 steps, with
validation every 1,000 steps; the depth-5 tree uses 128,000 steps,
with validation every 2,000 steps. Both protocols retain 64 validation
checkpoints. The nuisance-input settings use the corresponding training
budgets and validation schedules.

\subsection{Hyperparameter search and model selection}
\label{app:hpo}

Each method uses a fixed search space constructed on development targets
and frozen before final evaluation. A development panel selects the
validation-best configuration in each condition. For each searched
numerical parameter, the span of the selected values is expanded by
25\% of its width in log space, split equally between the two sides,
and clipped only to predefined domain boundaries.

For final evaluation, we use $T=32$ scrambled-Sobol configurations per
method, target instance, and training-set size. For each configuration,
we retain the checkpoint with lowest validation NMSE, then select the
configuration with lowest validation NMSE and evaluate it once on the
held-out test set. Table~\ref{tab:search-spaces} reports the frozen
search spaces.

\begin{table}[tbp]
\centering
\caption{
\textbf{Hyperparameter search spaces.}
Frozen ranges used in the final CSFRBench evaluation and reported
appendix experiments. All searched numerical parameters are sampled
log-uniformly. Search ranges are constructed on development targets
before final evaluation. The depth-5 tree uses separately frozen
ranges because it has a different backbone and training horizon.
A dash indicates that an experiment is not evaluated on that backbone.
}
\label{tab:search-spaces}
\small
\setlength{\tabcolsep}{3pt}
\begin{tabular}{llcc}
\toprule
Method & Parameter & $32\times4$ backbone & Depth-5 tree ($64\times8$) \\
\midrule
Dense & learning rate & $[0.00143,\,0.261]$ & $[7.70\times10^{-4},\,0.475]$ \\
$\ell_1$ & learning rate & $[1.57\times10^{-6},\,0.349]$ & $[2.72\times10^{-4},\,0.457]$ \\
 & $\lambda$ & $[10^{-7},\,670]$ & $[10^{-7},\,6.28\times10^{11}]$ \\
$\ell_2$ & learning rate & $[2.16\times10^{-5},\,0.648]$ & $[1.93\times10^{-5},\,0.115]$ \\
 & $\lambda$ & $[10^{-7},\,1.13\times10^{3}]$ & $[10^{-7},\,7.26\times10^{3}]$ \\
$L_0$ & learning rate & $[2.67\times10^{-4},\,0.0443]$ & $[2.46\times10^{-4},\,0.00839]$ \\
 & $\lambda$ & $[10^{-7},\,157]$ & $[10^{-7},\,75.5]$ \\
RigL & learning rate & $[1.01\times10^{-5},\,0.234]$ & $[1.00\times10^{-5},\,0.0121]$ \\
 & density & $[0.0309,\,1]$ & $[0.0155,\,0.720]$ \\
BIMT & learning rate & $[1.18\times10^{-5},\,0.125]$ & $[9.88\times10^{-7},\,0.265]$ \\
 & $\lambda$ & $[10^{-7},\,0.453]$ & $[10^{-7},\,1.35\times10^{3}]$ \\
Fiedler & learning rate & $[3.47\times10^{-5},\,0.268]$ & $[5.78\times10^{-6},\,0.0334]$ \\
 & $\delta$ & $[10^{-7},\,2.83\times10^{8}]$ & $[10^{-7},\,2.66\times10^{7}]$ \\
Neuron positions & learning rate & $[1.28\times10^{-5},\,0.431]$ & $[4.20\times10^{-7},\,0.194]$ \\
LeWi & learning rate & $[4.58\times10^{-6},\,0.0999]$ & $[7.92\times10^{-7},\,0.0712]$ \\
GraphOracle(1) & learning rate & $[0.00161,\,0.149]$ & $[5.28\times10^{-4},\,0.0301]$ \\
GraphOracle(2) & learning rate & $[0.00200,\,0.0799]$ & $[0.00102,\,0.0256]$ \\
GraphOracle(all) & learning rate & $[0.00200,\,0.0799]$ & $[0.00125,\,0.0190]$ \\
\midrule
\multicolumn{4}{l}{\emph{Appendix experiments (LeWi variants)}} \\
Mechanism ablations (3 arms) & learning rate & $[4.58\times10^{-6},\,0.0999]$ & -- \\
Mask-only forward & learning rate & $[4.58\times10^{-6},\,0.0999]$ & -- \\
$r \in \{1, 2, 4, 8\}$ & learning rate & $[4.58\times10^{-6},\,0.0999]$ & -- \\
Radius multiplier $\in \{0.5, 0.75, 1.5, 2\}$ & learning rate & $[4.58\times10^{-6},\,0.0999]$ & -- \\
Initialization scale $\in \{0.5, 2\}$ & learning rate & $[4.58\times10^{-6},\,0.0999]$ & -- \\
Seed replicates $\in \{1, 2, 3, 4\}$ & learning rate & $[4.58\times10^{-6},\,0.0999]$ & -- \\
\bottomrule
\end{tabular}
\par\smallskip\noindent\footnotesize The mechanism ablations are soft-only (no mask), fixed embeddings, and random sparse mask. Fixed settings: for $L_0$, $p_{\mathrm{open,init}} = 0.8318$; the GraphOracle levels and LeWi's $r = 3$, $R^2 = 1.5$ are method constants, not searched.
\end{table}

\paragraph{Search-space audit.}
As a post hoc check of the frozen search ranges, we measure the fraction
of selected hyperparameters lying within 10\% of either boundary in the
corresponding search coordinate. In the primary $d=8$ evaluation, across
all methods, searched parameters, and both backbones, the largest boundary
occupancy is $0.17$, and median normalized positions range from $0.31$ to
$0.76$. For LeWi, only $2\%$ of selected learning rates on the
$32\times4$ backbone and $3\%$ on the depth-5 backbone lie near a
boundary. Thus, the primary results show no substantial concentration of
selected hyperparameters at the edges of the frozen search spaces.
The nuisance-input evaluation shows the same pattern on the $32\times4$
backbone. 
On the depth-5 nuisance-input setting, the BIMT and $\ell_1$
regularization coefficients lie near a frozen search boundary for
$33\%$ and $26\%$ of selected configurations, respectively, so these
two baseline results may be sensitive to the chosen search range. No search space was modified based
on this audit.

\subsection{Statistical reporting}
\label{app:statistical-reporting}
Fixed-graph settings use 16 independently sampled target instances; the
sampled-graph setting uses 32 independently sampled graph/function
instances. We report arithmetic mean AULC with two-sided 95\%
Student-$t$ confidence intervals over instances. Direct method
comparisons use paired per-instance AULC differences with corresponding
paired Student-$t$ intervals.

Learning-curve figures show mean NMSE across instances, with
95\% Student-$t$ confidence intervals computed in log-NMSE space.
Confidence intervals are reported as uncertainty estimates for effect
sizes; comparisons whose paired interval contains zero are treated as
inconclusive rather than tied. We do not report $p$-values.

\section{Additional CSFRBench results and controls}
\label{app:additional-results}

\subsection{Complete predictive results}
\label{app:complete-predictive}

\begin{figure}[t]
    \centering
    \includegraphics[width=\linewidth]{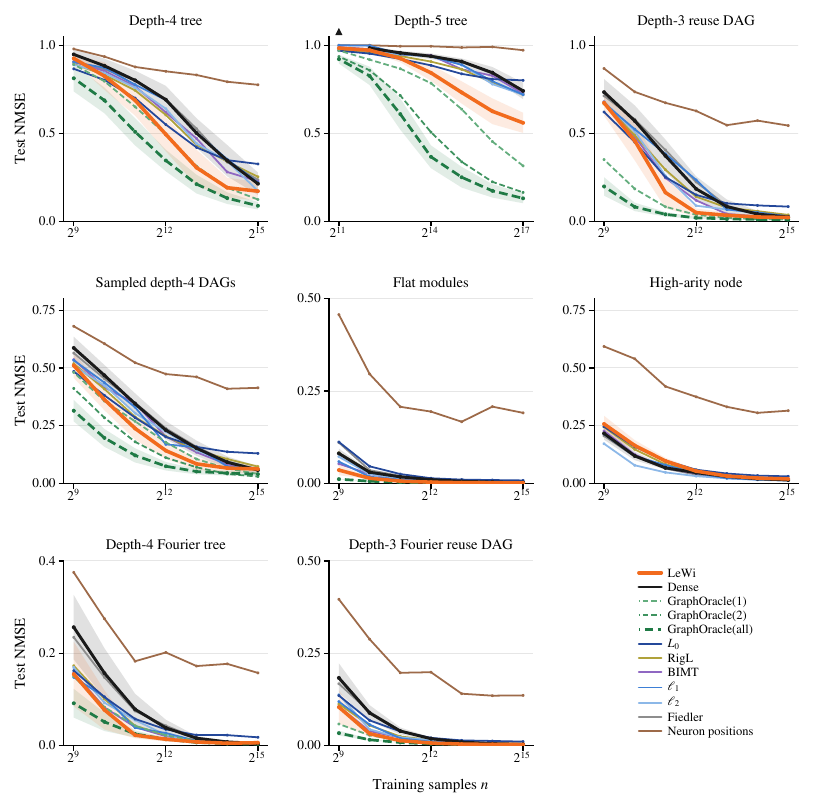}
    \caption{
\textbf{Complete CSFRBench learning curves.}
Held-out test NMSE versus training-set size for all evaluated methods and
settings. Fixed settings are averaged over 16 target instances and sampled
graphs over 32 graph/function instances. Bands show 95\% Student-$t$
confidence intervals for highlighted methods, and triangles mark means outside
the displayed range.
    }
    \label{fig:all-learning-curves}
\end{figure}

\begin{figure}[t]
    \centering
    \includegraphics[width=\linewidth]{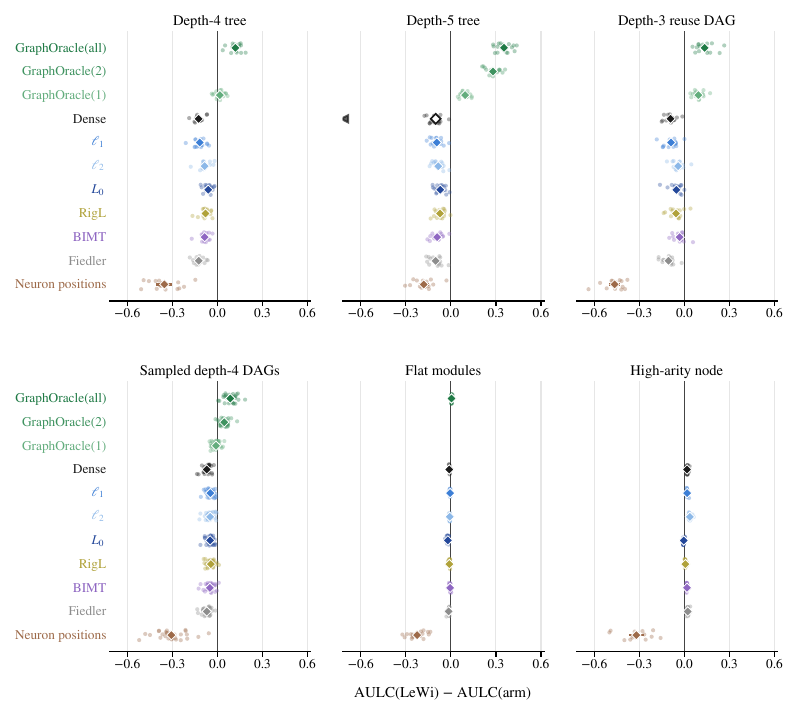}
    \caption{
\textbf{Paired predictive comparisons across CSFRBench.}
Paired AULC differences between LeWi and every baseline or graph-informed
reference. Negative values indicate lower AULC for LeWi. Points and intervals
show means and 95\% Student-$t$ confidence intervals over target instances.
Triangles mark means outside the displayed range, and hollow diamonds show
the corresponding medians.
    }
    \label{fig:all-aulc}
\end{figure}

Figure~\ref{fig:all-learning-curves} reports the complete learning
curves for all methods and CSFRBench settings.
Figure~\ref{fig:all-aulc} shows the corresponding paired AULC
comparisons with LeWi.
Across the low-arity settings, LeWi has lower mean AULC than every
learned baseline. The high-arity node provides the main exception,
where Dense achieves lower AULC than LeWi.
Tables~\ref{tab:aulc} and~\ref{tab:additional-structures} provide
the corresponding numerical results for the fixed settings, the
depth-5 tree, and independently sampled graphs.

\begin{table}[tbp]
\centering
\caption{
\textbf{Mean AULC on the fixed CSFRBench settings.}
Values are means over 16 target instances with 95\% Student-$t$ confidence
half-widths. AULC integrates test NMSE over
$n=2^9,\ldots,2^{15}$; lower is better.
}
\label{tab:aulc}
\small
\setlength{\tabcolsep}{2pt}
\begin{tabular}{lcccccc}
\toprule
 & Depth-4 & Depth-3 & Depth-4 & Flat & Fourier & High-arity \\
Method & tree & reuse DAG & Fourier tree & modules & reuse DAG & node \\
\midrule
LeWi & $0.508{\pm}0.062$ & $0.179{\pm}0.042$ & $0.035{\pm}0.013$ & $0.008{\pm}0.001$ & $0.019{\pm}0.006$ & $0.085{\pm}0.011$ \\
Dense & $0.633{\pm}0.059$ & $0.271{\pm}0.056$ & $0.071{\pm}0.023$ & $0.018{\pm}0.002$ & $0.043{\pm}0.009$ & $0.066{\pm}0.008$ \\
$\ell_1$ & $0.626{\pm}0.061$ & $0.268{\pm}0.061$ & $0.044{\pm}0.020$ & $0.013{\pm}0.002$ & $0.027{\pm}0.007$ & $0.066{\pm}0.008$ \\
$\ell_2$ & $0.593{\pm}0.061$ & $0.220{\pm}0.056$ & $0.046{\pm}0.019$ & $0.015{\pm}0.002$ & $0.027{\pm}0.005$ & $0.048{\pm}0.006$ \\
$L_0$ & $0.569{\pm}0.065$ & $0.232{\pm}0.060$ & $0.056{\pm}0.020$ & $0.027{\pm}0.004$ & $0.038{\pm}0.008$ & $0.089{\pm}0.010$ \\
RigL & $0.587{\pm}0.060$ & $0.234{\pm}0.060$ & $0.046{\pm}0.016$ & $0.017{\pm}0.003$ & $0.028{\pm}0.006$ & $0.077{\pm}0.010$ \\
BIMT & $0.592{\pm}0.061$ & $0.211{\pm}0.056$ & $0.040{\pm}0.015$ & $0.013{\pm}0.002$ & $0.023{\pm}0.005$ & $0.066{\pm}0.009$ \\
Fiedler & $0.632{\pm}0.054$ & $0.285{\pm}0.058$ & $0.068{\pm}0.023$ & $0.022{\pm}0.003$ & $0.042{\pm}0.009$ & $0.063{\pm}0.007$ \\
Neuron positions & $0.861{\pm}0.030$ & $0.643{\pm}0.059$ & $0.213{\pm}0.055$ & $0.233{\pm}0.032$ & $0.204{\pm}0.051$ & $0.404{\pm}0.059$ \\
GraphOracle(0) & -- & -- & -- & -- & -- & -- \\
GraphOracle(1) & $0.491{\pm}0.055$ & $0.085{\pm}0.026$ & $0.040{\pm}0.013$ & -- & $0.015{\pm}0.003$ & -- \\
GraphOracle(all) & $0.388{\pm}0.054$ & $0.044{\pm}0.012$ & $0.025{\pm}0.008$ & $0.003{\pm}0.000$ & $0.009{\pm}0.001$ & -- \\
\bottomrule
\end{tabular}
\end{table}

\begin{table}[tbp]
\centering
\caption{
\textbf{AULC on the depth-5 tree and on independently sampled depth-4 DAGs.}
Mean AULC with 95\% Student-$t$ half-widths, and
$\mathrm{AULC}(\mathrm{LeWi})-\mathrm{AULC}(\mathrm{method})$ as mean
[95\% paired Student-$t$ interval] over the number of instances on which LeWi has
lower AULC; negative values favor LeWi. The depth-5 tree (16 target instances;
width 64, eight hidden layers, 128{,}000 steps) integrates test NMSE over
$n=2^{11},\ldots,2^{17}$, the 32 independently sampled depth-4 DAG/function
instances over $n=2^9,\ldots,2^{15}$; neither range matches Table~\ref{tab:aulc}, so
AULC values are comparable within a column only.
}
\label{tab:additional-structures}
\small
\setlength{\tabcolsep}{2.5pt}
\begin{tabular}{lcccc}
\toprule
 & \multicolumn{2}{c}{Depth-5 tree} & \multicolumn{2}{c}{Independently sampled depth-4 DAGs} \\
\cmidrule(lr){2-3}\cmidrule(lr){4-5}
Method & AULC & LeWi $-$ method & AULC & LeWi $-$ method \\
\midrule
LeWi & $0.812{\pm}0.037$ & -- & $0.197{\pm}0.030$ & -- \\
Dense & $1.185{\pm}0.572$ & \begin{tabular}[t]{@{}c@{}}$-0.374$ [$-0.950$, $+0.203$]\\(16/16)\end{tabular} & $0.268{\pm}0.033$ & \begin{tabular}[t]{@{}c@{}}$-0.072$ [$-0.080$, $-0.063$]\\(32/32)\end{tabular} \\
$\ell_1$ & $0.905{\pm}0.020$ & \begin{tabular}[t]{@{}c@{}}$-0.093$ [$-0.114$, $-0.073$]\\(16/16)\end{tabular} & $0.244{\pm}0.031$ & \begin{tabular}[t]{@{}c@{}}$-0.047$ [$-0.056$, $-0.038$]\\(32/32)\end{tabular} \\
$\ell_2$ & $0.895{\pm}0.021$ & \begin{tabular}[t]{@{}c@{}}$-0.084$ [$-0.103$, $-0.065$]\\(16/16)\end{tabular} & $0.245{\pm}0.033$ & \begin{tabular}[t]{@{}c@{}}$-0.049$ [$-0.059$, $-0.039$]\\(32/32)\end{tabular} \\
$L_0$ & $0.882{\pm}0.025$ & \begin{tabular}[t]{@{}c@{}}$-0.070$ [$-0.089$, $-0.052$]\\(16/16)\end{tabular} & $0.245{\pm}0.033$ & \begin{tabular}[t]{@{}c@{}}$-0.048$ [$-0.056$, $-0.040$]\\(32/32)\end{tabular} \\
RigL & $0.884{\pm}0.023$ & \begin{tabular}[t]{@{}c@{}}$-0.072$ [$-0.090$, $-0.054$]\\(16/16)\end{tabular} & $0.241{\pm}0.032$ & \begin{tabular}[t]{@{}c@{}}$-0.044$ [$-0.053$, $-0.035$]\\(31/32)\end{tabular} \\
BIMT & $0.903{\pm}0.020$ & \begin{tabular}[t]{@{}c@{}}$-0.091$ [$-0.111$, $-0.071$]\\(16/16)\end{tabular} & $0.247{\pm}0.033$ & \begin{tabular}[t]{@{}c@{}}$-0.050$ [$-0.061$, $-0.040$]\\(31/32)\end{tabular} \\
Fiedler & $0.913{\pm}0.018$ & \begin{tabular}[t]{@{}c@{}}$-0.102$ [$-0.124$, $-0.080$]\\(16/16)\end{tabular} & $0.268{\pm}0.033$ & \begin{tabular}[t]{@{}c@{}}$-0.071$ [$-0.081$, $-0.062$]\\(32/32)\end{tabular} \\
Neuron positions & $0.992{\pm}0.003$ & \begin{tabular}[t]{@{}c@{}}$-0.181$ [$-0.215$, $-0.146$]\\(16/16)\end{tabular} & $0.503{\pm}0.037$ & \begin{tabular}[t]{@{}c@{}}$-0.307$ [$-0.340$, $-0.273$]\\(32/32)\end{tabular} \\
GraphOracle(0) & -- & -- & -- & -- \\
GraphOracle(1) & $0.717{\pm}0.038$ & \begin{tabular}[t]{@{}c@{}}$+0.095$ [$+0.079$, $+0.110$]\\(0/16)\end{tabular} & $0.206{\pm}0.030$ & \begin{tabular}[t]{@{}c@{}}$-0.009$ [$-0.018$, $+0.000$]\\(20/32)\end{tabular} \\
GraphOracle(2) & $0.531{\pm}0.047$ & \begin{tabular}[t]{@{}c@{}}$+0.280$ [$+0.255$, $+0.305$]\\(0/16)\end{tabular} & $0.151{\pm}0.024$ & \begin{tabular}[t]{@{}c@{}}$+0.046$ [$+0.035$, $+0.056$]\\(1/32)\end{tabular} \\
GraphOracle(all) & $0.458{\pm}0.051$ & \begin{tabular}[t]{@{}c@{}}$+0.353$ [$+0.326$, $+0.381$]\\(0/16)\end{tabular} & $0.111{\pm}0.022$ & \begin{tabular}[t]{@{}c@{}}$+0.086$ [$+0.072$, $+0.100$]\\(0/32)\end{tabular} \\
\bottomrule
\end{tabular}
\end{table}

\subsection{Representation diagnostics}
\label{app:representation-diagnostics}

\begin{figure}[t]
    \centering
    \includegraphics[width=\linewidth]
    {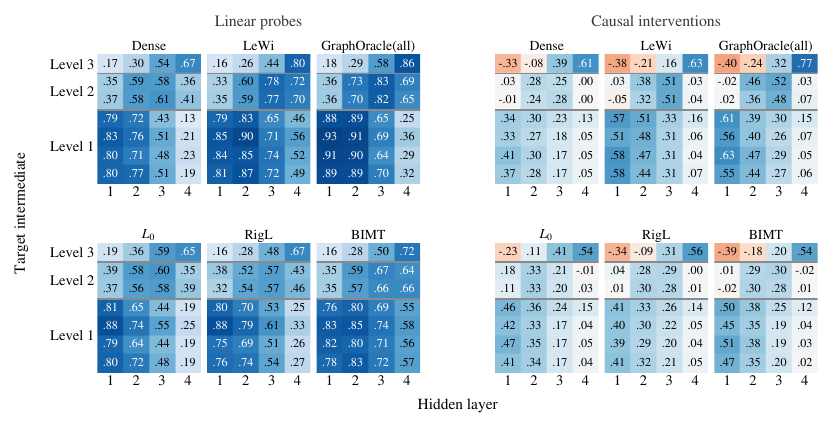}
    \caption{
        \textbf{Representation diagnostics for additional learned baselines.}
        Left: held-out $R^2$ from ridge probes predicting each target
        intermediate from each hidden layer. Right: held-out counterfactual
        $R^2$ under joint multi-variable Boundless DAS. The top row repeats
        Dense, LeWi, and GraphOracle(all) from
        Figure~\ref{fig:learned-structure}, while the bottom row shows
        $L_0$, RigL, and BIMT. Rows are grouped by target level and columns
        correspond to hidden layers. Results are for the depth-4 tree at
        $n=2^{14}$, averaged over 16 target instances.
    }
    \label{fig:representation-baselines}
\end{figure}

\paragraph{Additional learned baselines.}
Figure~\ref{fig:representation-baselines} extends the main
representation analysis to $L_0$, RigL, and BIMT. LeWi generally
shows higher counterfactual agreement than these learned baselines
for level-1 and level-2 intermediates. BIMT also exhibits relatively
high linear decodability without corresponding intervention scores,
further illustrating that decodability and use in prediction are
distinct diagnostics.

\begin{figure}[t]
    \centering
    \includegraphics[width=\linewidth]
    {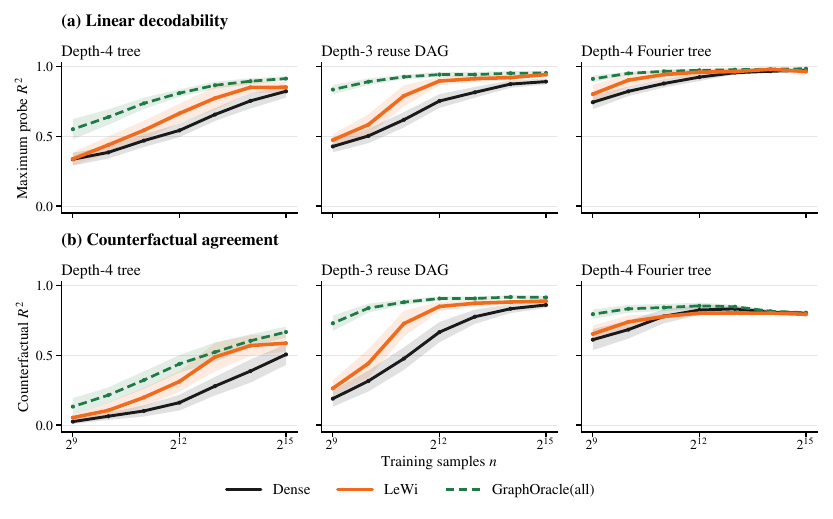}
    \caption{
        \textbf{Representation diagnostics across training-set size and target structure.}
        (a) Maximum held-out ridge-probe $R^2$ on the depth-4 tree,
        reuse DAG, and Fourier tree. For each target intermediate,
        we take the maximum $R^2$ across hidden layers and then average
        across target intermediates and target instances.
        (b) Mean held-out counterfactual $R^2$ under joint multi-variable
        Boundless DAS on the same target families.
        Curves show Dense, LeWi, and GraphOracle(all), averaged over
        16 target instances with 95\% Student-$t$ confidence intervals.
    }
    \label{fig:representation-sample-size}
\end{figure}

\paragraph{Across training-set size and target structure.}
Figure~\ref{fig:representation-sample-size} evaluates both diagnostics
across training-set sizes on the depth-4 tree, reuse DAG, and Fourier
tree. LeWi generally exceeds Dense in both maximum probe $R^2$ and
counterfactual $R^2$ on the tree and reuse DAG. The differences are
smaller on the Fourier tree, where the methods converge at larger
training-set sizes. These results show that the representation
differences in Figure~\ref{fig:learned-structure} are not specific
to a single training-set size or target family.

\subsection{Nuisance-input results}
\label{app:nuisance-input-results}

We repeat the benchmark on the same target instances after increasing
the ambient dimension to $d=32$. The eight-input settings receive
24 inert coordinates, while the depth-5 tree receives 16 inert
coordinates in addition to its 16 active inputs. All other elements
of the training and evaluation protocol are unchanged. Thus, the
all-active and nuisance-input settings can be compared on matched
target instances.

\begin{figure}[t]
    \centering
    \includegraphics[width=\linewidth]{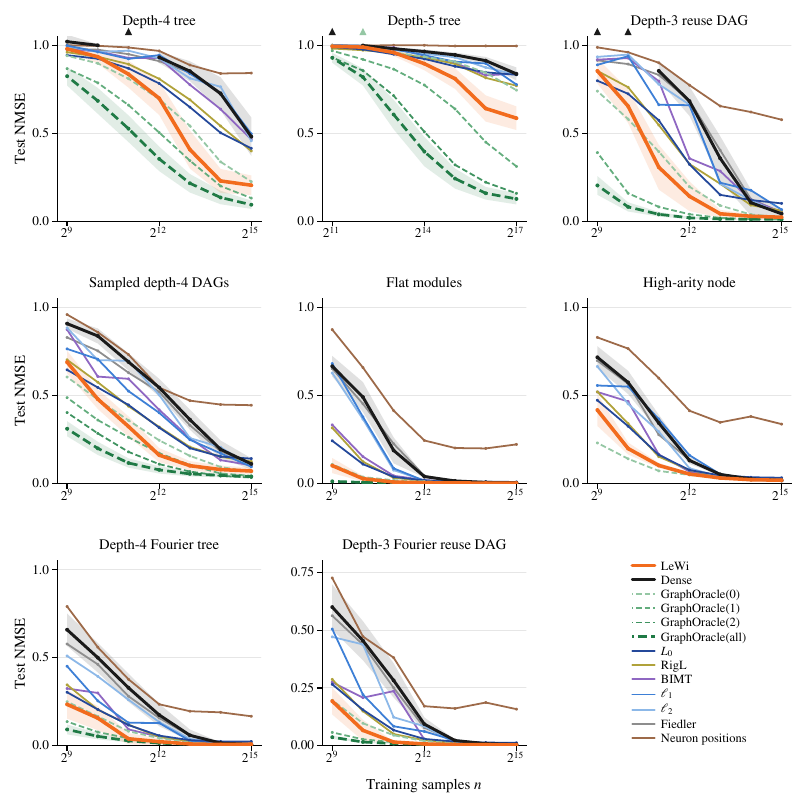}
    \caption{
\textbf{Complete CSFRBench learning curves with nuisance inputs.}
Held-out test NMSE versus training-set size with ambient dimension $d=32$.
The depth-4 settings contain eight active and 24 inert inputs, while the
depth-5 tree contains 16 active and 16 inert inputs. Fixed settings are
averaged over 16 target instances and sampled graphs over 32 graph/function
instances. Bands show 95\% Student-$t$ confidence intervals for highlighted
methods, and triangles mark means outside the displayed range.
    }
    \label{fig:nuisance-learning-curves}
\end{figure}

\begin{figure}[t]
    \centering
    \includegraphics[width=\linewidth]{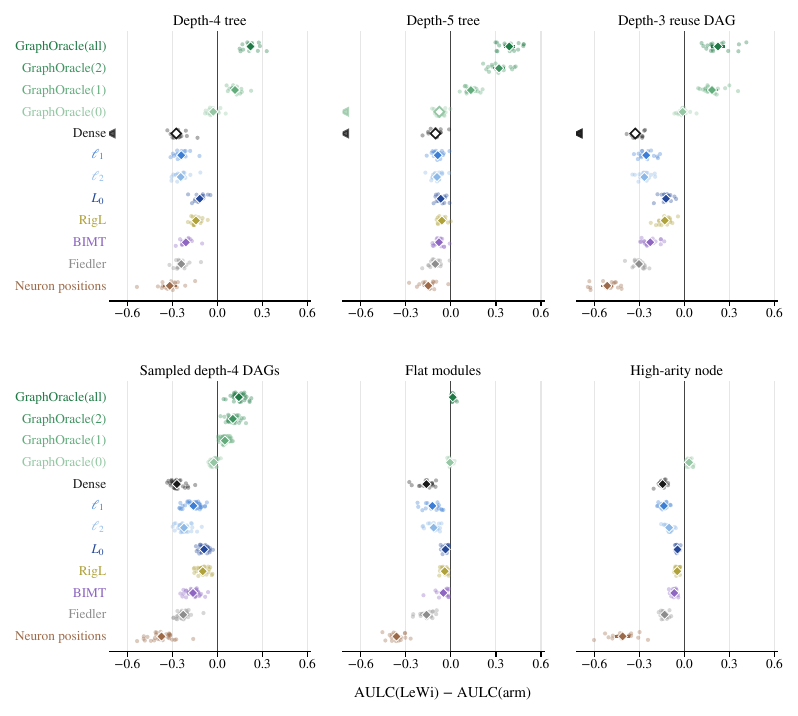}
    \caption{
\textbf{Paired predictive comparisons with nuisance inputs.}
Paired AULC differences between LeWi and every baseline or graph-informed
reference at $d=32$. Negative values indicate lower AULC for LeWi. Points and
intervals show means and 95\% Student-$t$ confidence intervals over target
instances. Triangles mark means outside the displayed range, and hollow
diamonds show the corresponding medians.
    }
    \label{fig:nuisance-aulc}
\end{figure}

Figure~\ref{fig:nuisance-learning-curves} reports the complete
nuisance-input learning curves, including the depth-5 tree.
Figure~\ref{fig:nuisance-aulc} gives the corresponding paired
comparisons against the learned baselines and graph-informed
references.

\subsection{Wiring alignment}
\label{app:structure-discovery}

\paragraph{Module-graph extraction.}
For each trained LeWi network, we retain hidden neurons lying on an
active input-to-output path and cluster their embeddings separately
within each hidden layer by average linkage, cutting the dendrogram at
the connection radius $R$. Each input and the output form individual
modules. We connect two modules in adjacent layers when more than half
of their candidate neuron-to-neuron connections are active. This step
uses only the learned embeddings and masks.

\paragraph{Sensitivity to module extraction.}
The reported analysis uses average-linkage clustering with a dendrogram cut
at $R$. On development targets, we also evaluated cuts at $0.75R$ and
$1.25R$ and alternative linkage rules. The resulting scores depend on the
extraction rule, so we fix average linkage at $R$ for the reported evaluation.

\paragraph{Target matching and scoring.}
We then use target intermediates only to assign semantic
correspondences. For each candidate module and target intermediate,
we fit a ridge probe and compare its validation $R^2$ with 200 random
neuron sets matched in size and layer. Pairs above the 95th percentile
of these reference scores are eligible for a one-to-one assignment
chosen to maximize validation $R^2$. Dependencies are inferred from
paths between matched modules, including paths through unmatched
modules, and the resulting graph is transitively reduced. We report
edge precision, recall, and F1 over all target edges and separately
over edges originating at target intermediate nodes. Because target
intermediates are used for post hoc matching, these metrics measure
structural agreement with the target graph rather than graph recovery.

\begin{figure}[t]
    \centering
    \includegraphics[width=\linewidth]
    {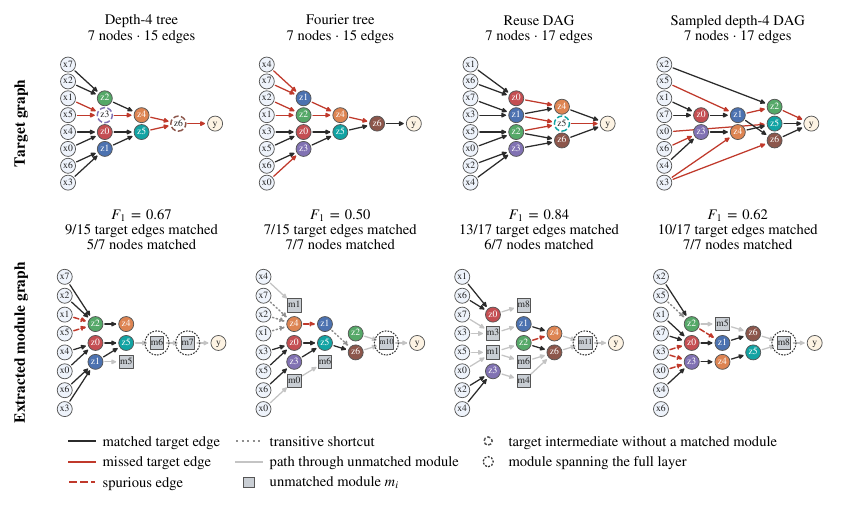}
    \caption{
\textbf{Representative target and extracted module graphs.}
Each column shows a target computation graph (top) and the module graph
extracted from one trained LeWi network (bottom). Colors denote post hoc
matches between modules and target intermediates. Black edges correspond to
target dependencies, red target edges are missed, and the remaining edge
styles indicate spurious connections, transitive shortcuts, or paths through
unmatched modules. These examples illustrate the graph comparison summarized
by edge precision, recall, and $F_1$.
Each example is the network with upper-median edge $F_1$ in its setting.
    }
    \label{fig:module-graphs}
\end{figure}

\begin{figure}[t]
    \centering
    \includegraphics[width=\linewidth]
    {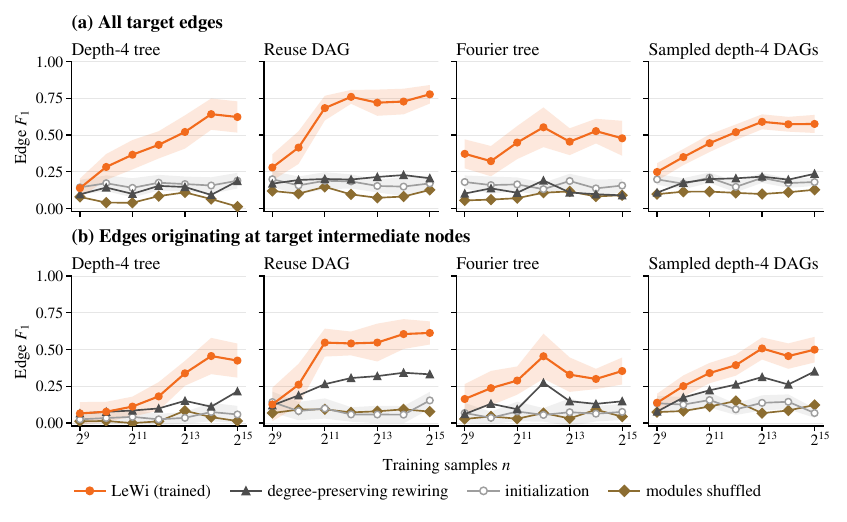}
    \caption{
        \textbf{Wiring alignment across training-set size.}
        Edge F1 of the extracted module graph against the target computation
        graph as the number of training samples increases.
        Filled markers report F1 over all target edges; hollow markers report
        F1 for edges originating at target intermediate nodes.
        Curves show trained LeWi and the corresponding initialization and
        randomized controls, averaged over target instances with 95\%
        Student-$t$ confidence intervals.
    }
    \label{fig:wiring-vs-n}
\end{figure}

\paragraph{Results.}
Table~\ref{tab:wiring-alignment} reports per-setting precision, recall,
and $F_1$. Precision exceeds recall in every setting, so missed target
edges are more common than spurious edges. Some missed edges arise when
a target intermediate has no matched module. Figure~\ref{fig:module-graphs}
shows representative target and extracted module graphs, and
Figure~\ref{fig:wiring-vs-n} shows alignment across training-set size.

\begin{figure}[p]
    \centering
    \includegraphics[width=\linewidth]{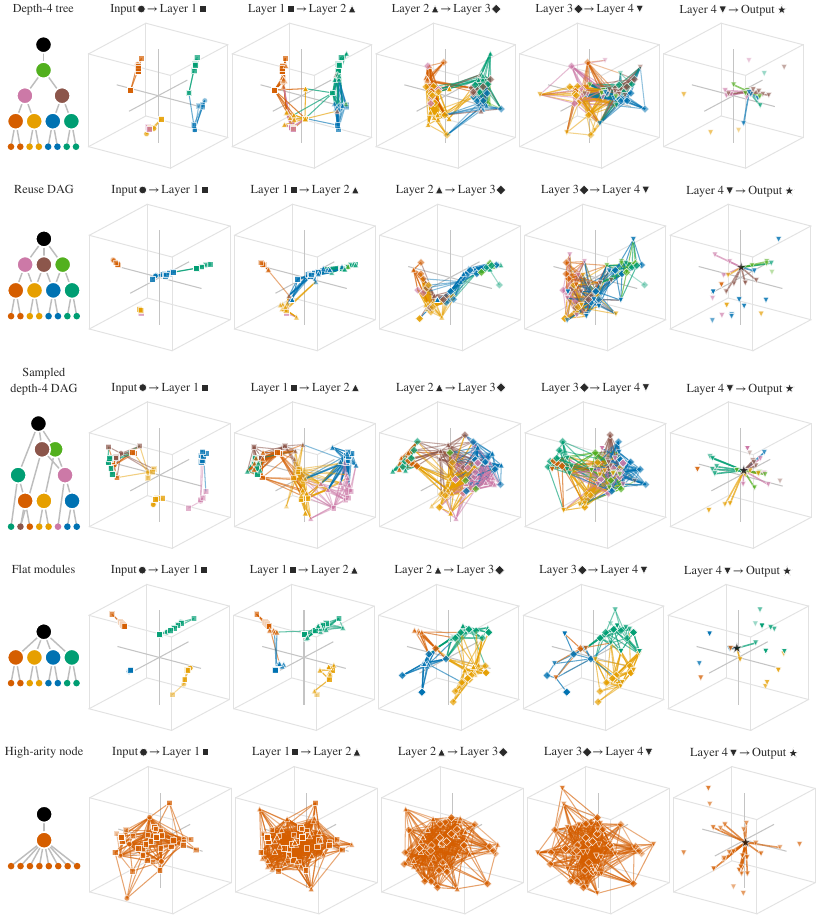}
    \caption{
\textbf{Learned neuron embeddings across target structures.}
Examples from trained LeWi models on the depth-4 tree, reuse DAG,
flat modules, high-arity node, and a sampled depth-4 DAG.
Each row shows the target computation graph followed by the learned
three-dimensional neuron embeddings across successive network-layer
transitions. Lines denote active connections, and colors indicate post hoc
associations with target intermediates. Additional embedding visualizations
and their training trajectories are available in the interactive demo linked
in Figure~\ref{fig:teaser}.
    }
    \label{fig:embedding-geometry}
\end{figure}

\paragraph{Embedding visualizations.}
The low-dimensional neuron embeddings also provide a direct view of the
learned wiring. Figure~\ref{fig:embedding-geometry} extends the visualization
in Figure~\ref{fig:teaser} across several target structures. Colors indicate
post hoc associations with target intermediates, and lines indicate active
connections. These embedding visualizations and their training trajectories
are also available in the interactive demo linked in Figure~\ref{fig:teaser}.

\begin{table}[tbp]
\centering
\caption{
\textbf{Wiring alignment at $n=2^{15}$.}
Precision, recall, and $F_1$ between the extracted module graph and the
target computation graph, reported over all target edges and over edges
originating at target intermediate nodes. Assessable recall restricts to
target edges whose endpoints have matched modules.
}
\label{tab:wiring-alignment}
\small
\setlength{\tabcolsep}{4.5pt}
\begin{tabular}{lcccccccc}
\toprule
Setting & \multicolumn{3}{c}{All edges} & \multicolumn{3}{c}{From intermediates} & \multicolumn{2}{c}{Assessable edges} \\
\cmidrule(lr){2-4}\cmidrule(lr){5-7}\cmidrule(lr){8-9}
 & Precision & Recall & $F_1$ & Precision & Recall & $F_1$ & Recall & Share \\
\midrule
Depth-4 tree & $0.677$ & $0.579$ & $0.623$ & $0.503$ & $0.375$ & $0.425$ & $0.731$ & $0.800$ \\
Reuse DAG & $0.886$ & $0.695$ & $0.777$ & $0.829$ & $0.493$ & $0.613$ & $0.772$ & $0.915$ \\
Fourier tree & $0.518$ & $0.446$ & $0.478$ & $0.403$ & $0.321$ & $0.354$ & $0.508$ & $0.895$ \\
Sampled depth-4 DAGs & $0.674$ & $0.509$ & $0.575$ & $0.634$ & $0.423$ & $0.499$ & $0.542$ & $0.935$ \\
\bottomrule
\end{tabular}
\par\smallskip\noindent\footnotesize Values are means of instance-level metrics rather than metrics computed
from pooled edge counts.
\end{table}

\subsection{Learned connection density}
\label{app:connection-density}

\begin{figure}[t]
    \centering
    \includegraphics[width=\linewidth]{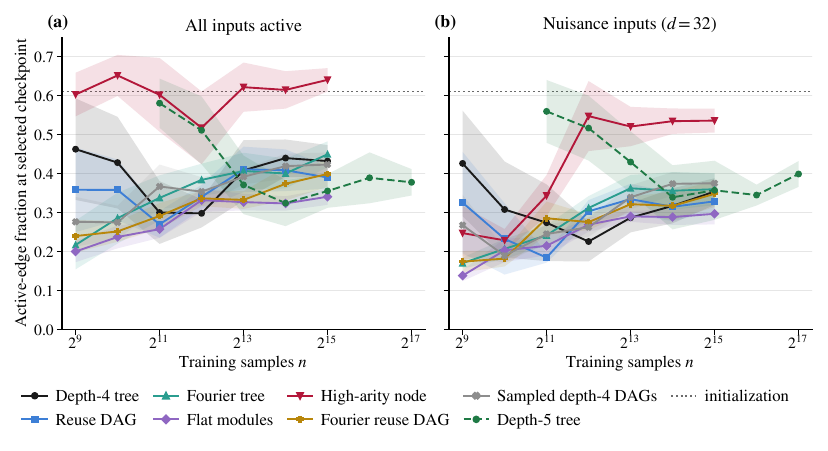}
    \caption{
\textbf{Learned connection density.}
Active-edge fraction at the selected checkpoint across training-set size
with (a) all inputs active and (b) nuisance inputs at $d=32$.
The dotted line marks the initialization density of $0.61$.
The depth-5 tree uses $d=16$ in (a) and 16 active plus 16 nuisance
inputs in (b); the other settings use 8 active inputs and, in (b),
24 nuisance inputs. Curves show means over target instances with
95\% Student-$t$ confidence intervals.
    }
    \label{fig:connection-density}
\end{figure}

Figure~\ref{fig:connection-density} reports the fraction of active
connections in the selected LeWi models across training-set size.
Initialization has active-edge fraction $0.61$ in every setting.
On the all-active compositional settings, the selected models have
densities of $0.29$--$0.40$, while the high-arity control remains near
its initialization density. With nuisance inputs, the corresponding
compositional settings fall further to $0.24$--$0.31$, while the
high-arity control retains density $0.42$. The depth-5 tree is near
$0.42$ in both conditions.

The sparsity is also structured across network depth. The input
transition is generally the sparsest, reaching active-edge fractions of
$0.07$--$0.29$ with nuisance inputs, while the readout is typically the
densest transition on compositional targets. Density increases modestly
with training-set size in most settings. The depth-5 tree is the main
exception, decreasing from $0.58$ at $n=2^{11}$ to $0.38$ at
$n=2^{17}$ in the all-active setting, with a similar trend when nuisance
inputs are present.

\subsection{Parameterization sensitivity}
\label{app:parameterization-analysis}

\begin{figure}[t]
    \centering
    \includegraphics[width=\linewidth]
    {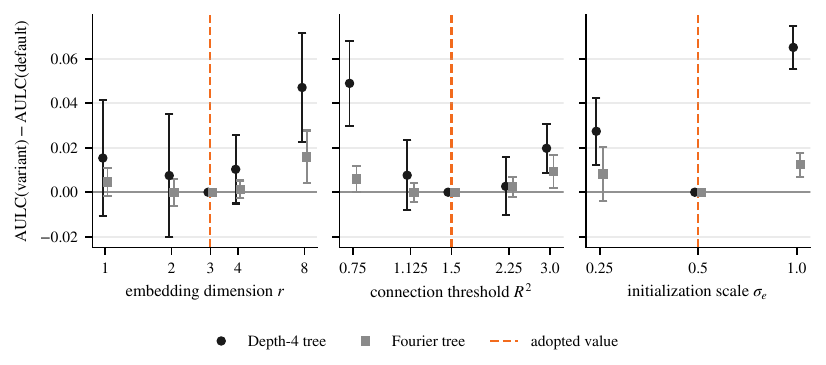}
    \caption{
\textbf{Sensitivity to LeWi's geometric parameterization.}
Paired AULC differences from the adopted configuration when varying
the embedding dimension $r$ (left), connection threshold $R^2$ with
$\sigma_e$ fixed (middle), and initialization scale $\sigma_e$ with
$R^2=2r\sigma_e^2$ (right).
Positive values indicate worse sample efficiency than the adopted
configuration. Results are shown for the depth-4 tree and Fourier tree.
Error bars show 95\% Student-$t$ confidence intervals over target
instances. Dashed lines mark the adopted values
$r=3$, $R^2=1.5$, and $\sigma_e=0.5$.
    }
    \label{fig:parameterization-sensitivity}
\end{figure}

We test sensitivity to the embedding dimension $r$, connection
threshold $R^2$, and initialization scale $\sigma_e$
(Figure~\ref{fig:parameterization-sensitivity}). Performance is
relatively stable near the adopted configuration. Embedding dimensions
$r\in\{1,2,3,4\}$ perform similarly, while $r=8$ increases AULC.
Moderate changes to $R^2$ also have small effects, whereas larger
departures can reduce sample efficiency. Sensitivity to $\sigma_e$ is
stronger on the depth-4 tree, particularly at the larger initialization
scale. Overall, the adopted configuration lies in a relatively stable
region rather than at a sharply isolated optimum.

\subsection{Computational cost}
\label{app:computational-cost}

LeWi adds 5--12\% more parameters than Dense. Its computational
overhead comes from computing and differentiating pairwise embedding
distances, gates, and masks. We accelerate training by batching
independent networks and using fused C++ kernels, while similarly
optimizing Dense through batching and compilation. These optimizations
preserve the original training algorithms and protocols.
In matched CPU experiments, LeWi incurs 32\% and 2\% higher
end-to-end runtime than Dense on the $32\times4$ and $64\times8$
backbones, respectively (Table~\ref{tab:overhead-compact}).
LeWi still evaluates all candidate connections densely, so learned
sparsity does not reduce training computation. At inference,
the learned gates and masks can be folded into the weights.

\begin{table}[tbp]
\centering
\caption{
\textbf{Computational cost of LeWi and Dense.}
Parameters and runtime (seconds per trained network) using
optimized CPU implementations. Runtime includes training,
validation, and testing, with both methods run concurrently
on the same node (one thread per process).
}
\label{tab:overhead-compact}
\footnotesize
\setlength{\tabcolsep}{5pt}
\begin{tabular}{lrrrrr}
\toprule
& \multicolumn{2}{c}{Parameters}
& \multicolumn{3}{c}{Runtime (s)} \\
\cmidrule(lr){2-3}
\cmidrule(lr){4-6}
Backbone & Dense & LeWi & Dense & LeWi & Ratio \\
\midrule
$32\times4$
& 3{,}489 & 3{,}900 & 5.2 & 6.8 & 1.32 \\
$64\times8$
& 30{,}273 & 31{,}860 & 84.3 & 85.6 & 1.02 \\
\bottomrule
\end{tabular}
\end{table}

\section{World-model experiments}
\label{app:world-models}

\subsection{Task suite}
\label{app:world-model-tasks}

We use the state-based, continuous-action environments of
\texttt{stable-worldmodel}~\citep{maes2026stable}. The registry contains
21 environments that run under this interface. At the final $256\times4$
protocol, 17 satisfy the fixed eligibility criteria described below and
form the evaluation suite.

The tasks span state dimensions from 2 to 78 and action dimensions from
1 to 12. They cover qualitatively different transition structure:
Pendulum and Cartpole have smooth low-order dynamics; Cheetah, Hopper,
Walker, and Quadruped contain articulated rigid-body dynamics with local
joint interactions; Ball in cup and the manipulation tasks introduce
contact-dependent transitions; the navigation environments combine
simple motion with wall constraints; and Piecewise uses a
region-dependent transition law. Thus the suite varies both ambient
dimension and the form of structure available in the dynamics.

\paragraph{Task eligibility.}
The screen retains tasks on which predictive learning methods can be
meaningfully distinguished. It uses only Dense, ridge regression, a persistence
predictor, and simulator replay; LeWi and the regularized baselines do not enter
task selection. A task must improve over persistence and ridge regression,
avoid numerical floor and ceiling regimes, not be dominated by model capacity,
exhibit a data-limited regime over the evaluated sample sizes, and replay
correctly in the simulator. The criteria and thresholds are fixed across tasks.
Under the width-256, four-hidden-layer protocol, 17 of the 21 registry tasks
satisfy all criteria and form the evaluation suite. Humanoid walk and OGBench
cube were run under the full protocol but failed the screen, while Mountain car
and OGBench scene failed before full evaluation. Humanoid walk failed the
Dense-versus-ridge criterion; OGBench cube additionally lacked a data-limited
regime and reliable simulator replay; Mountain car lacked a data-limited
regime; and OGBench scene failed the Dense-versus-ridge and data-limited-regime
criteria.

\subsection{Experimental protocol}
\label{app:world-model-protocol}

\paragraph{Data.}
For each task, we collect exploratory trajectories using temporally
correlated random actions within the environment's action bounds.
Datasets are split by episode into 80\% training, 10\% validation, and
10\% test sets. Training sets are nested prefixes with
$n\in\{2^{13},\ldots,2^{19}\}$ transitions and are shared across
methods. Angular coordinates are represented by sine and cosine pairs.
Unit-circle and quaternion features are renormalized after each predicted
rollout step.

\paragraph{Prediction problem.}
Models receive the current state and action and predict the change in
state. State changes are standardized using statistics computed from the
corresponding training subset. At evaluation time, predictions are
rolled forward autoregressively using the recorded action sequence.

\paragraph{Training and model selection.}
All methods use a width-256, four-hidden-layer ReLU MLP with vector output.
Models are trained for 128,000 Adam steps with batch size 32 and cosine
learning-rate decay to 1\% of the initial rate. Validation is performed every
500 steps and the best validation checkpoint is retained. For each task,
training-set size, and method, we evaluate 32 scrambled-Sobol hyperparameter
configurations, ranked by mean validation rollout NMSE over horizons
$1,2,4,8,16$, with one-step validation NMSE used only as a tie-breaker.
The selected configuration is evaluated using 16 paired final seeds.
Table~\ref{tab:world-model-search-spaces} reports the final search spaces.
For Dense and $\ell_1$, a post hoc audit found selected learning rates within
10\% of the initial lower bound in 32\% and 25\% of task--training-set-size
combinations, respectively. Before rerunning either method, we fixed the
correction rule to extend only its lower learning-rate bound by one decade,
leaving all other search ranges unchanged. We then repeated the full
32-configuration search and final evaluation for both methods.
LeWi uses the same initialization checks as in CSFRBench
(Appendix~\ref{app:lewi-implementation}).

\begin{table}[tbp]
\centering
\caption{\textbf{Hyperparameter search spaces for the world-model experiments.}
All searched numerical parameters are sampled log-uniformly.}
\label{tab:world-model-search-spaces}
\small
\setlength{\tabcolsep}{4pt}
\begin{tabular}{llc}
\toprule
Method & Parameter & Search range \\
\midrule
Dense & learning rate & $[8.58\times10^{-5},\,4.63\times10^{-2}]$ \\
$\ell_1$ & learning rate & $[4.94\times10^{-5},\,9.85\times10^{-2}]$ \\
 & $\lambda$ & $[1.66\times10^{-9},\,1.66\times10^{3}]$ \\
$\ell_2$ & learning rate & $[3.44\times10^{-4},\,3.98\times10^{-2}]$ \\
 & $\lambda$ & $[5.22\times10^{-7},\,9.06\times10^{1}]$ \\
$L_0$ & learning rate & $[1.30\times10^{-3},\,6.50\times10^{-2}]$ \\
 & $\lambda$ & $[2.75\times10^{-9},\,1.61\times10^{2}]$ \\
RigL & learning rate & $[1.41\times10^{-4},\,1.60\times10^{-2}]$ \\
 & density & $[1.58\times10^{-2},\,1]$ \\
BIMT & learning rate & $[3.85\times10^{-4},\,1.22\times10^{-2}]$ \\
 & $\lambda$ & $[1.25\times10^{-8},\,7.36\times10^{-3}]$ \\
Fiedler & learning rate & $[1.83\times10^{-4},\,1.68\times10^{-2}]$ \\
 & $\delta$ & $[2.03\times10^{-9},\,2.84\times10^{5}]$ \\
Learned neuron positions & learning rate & $[2.15\times10^{-4},\,1.35\times10^{-1}]$ \\
LeWi & learning rate & $[9.66\times10^{-5},\,2.53\times10^{-2}]$ \\
\bottomrule
\end{tabular}
\end{table}

\paragraph{Validation-only stability re-selection.}
To reduce sensitivity to unstable configurations selected from a single search
seed, we apply a second validation-only selection stage to cells satisfying a
fixed instability criterion. A cell is flagged when the geometric mean
validation rollout error across its final seeds is at least $1.5\times$ that of
the selected search trial, or an individual final exceeds $10\times$ the
cell median, and its mean validation log-error exceeds both neighboring
training-set sizes by at least $0.10$. For a flagged cell, the four
highest-ranked first-stage configurations are evaluated on eight fresh paired
seeds and re-ranked by mean $\log_{10}$ validation rollout NMSE, with one-step
NMSE and the original rank as tie-breakers. If the selected configuration
changes, it is retrained on the same 16 paired final seeds. Test values never
enter the trigger or either selection stage.

\paragraph{Evaluation.}
Our primary metric is open-loop NMSE at horizon 16, averaged over the
informative state dimensions. For each final seed, we summarize sample
efficiency as $\mathrm{AULC}_{\mathrm{WM}}$, the trapezoidal area under
$\log_{10}$ rollout-16 NMSE as a function of $\log_2 n$. Method comparisons
use paired per-seed AULC differences with two-sided 95\% Student-$t$
confidence intervals. A task is counted as a win or loss when this interval
excludes zero and otherwise as a tie. An AULC difference $\Delta$ corresponds
to a multiplicative factor $10^\Delta$ in rollout NMSE averaged geometrically
over the evaluated sample sizes.

\subsection{Complete results}
\label{app:world-model-results}

\begin{figure*}[t]
    \centering
    \includegraphics[width=\textwidth]{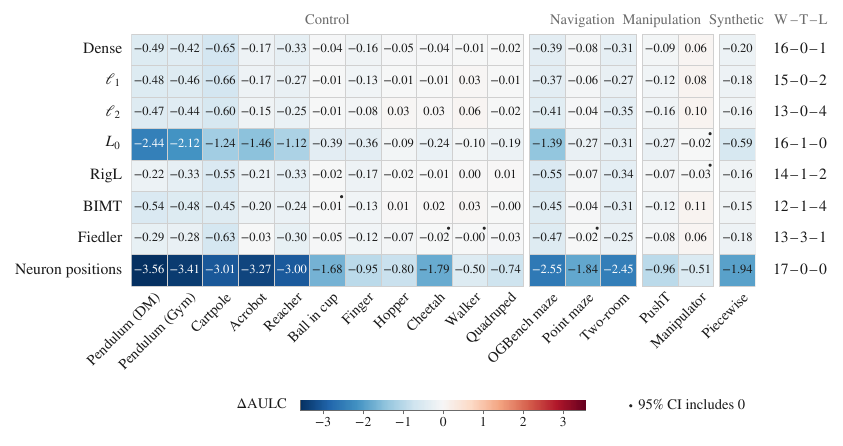}
    \caption{
    \textbf{Complete world-model comparisons.}
    Each cell shows the mean paired
    $\Delta\mathrm{AULC}
    =\mathrm{AULC}(\mathrm{LeWi})-\mathrm{AULC}(\mathrm{baseline})$
    over 16 paired seeds; negative values favor LeWi.
    A dot marks a comparison whose 95\% Student-$t$ confidence interval contains zero. The right margin reports LeWi's win--tie--loss count against each
    baseline across the 17 tasks.
    }
    \label{fig:world-model-full}
\end{figure*}

\begin{figure*}[t]
    \centering
    \includegraphics[width=\textwidth]{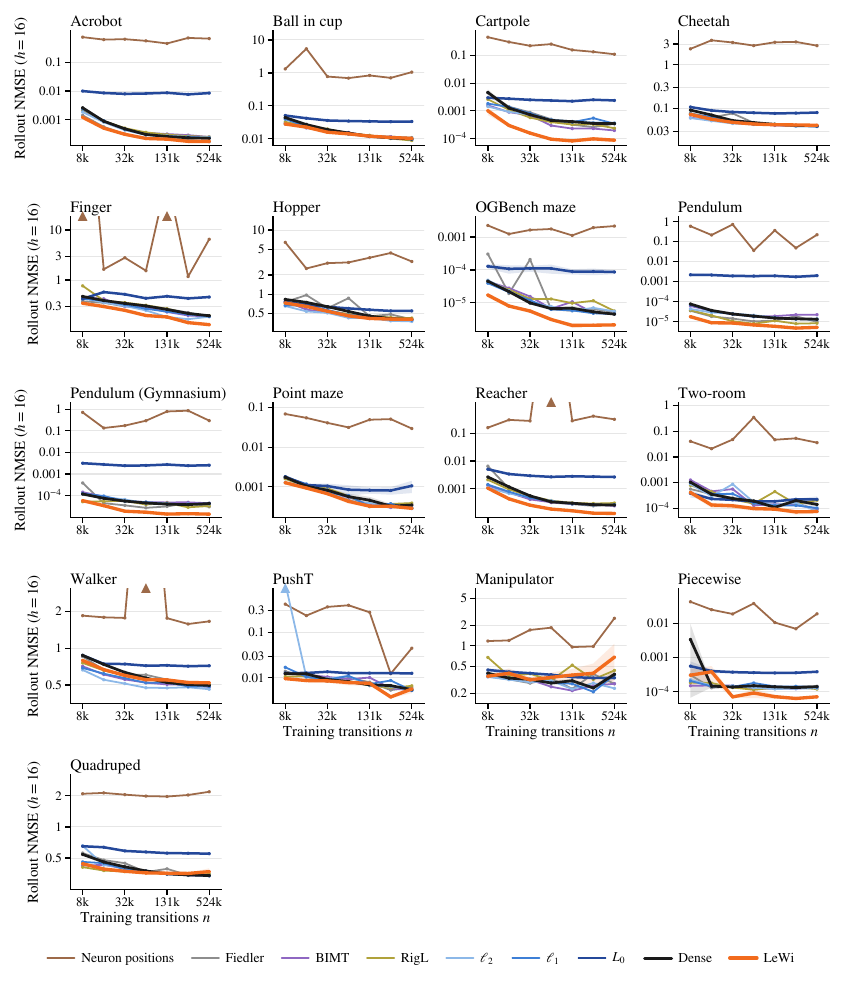}
    \caption{
    \textbf{Complete world-model learning curves.}
    Rollout-16 NMSE versus the number of training transitions for all 17 tasks
    and nine methods. Curves show arithmetic means over 16 paired final seeds
    with 95\% Student-$t$ confidence intervals; lower is better.
    Triangles indicate means outside the displayed range.
    Learned neuron positions are included here but omitted from the main-text
    heatmap because their larger errors would compress its color scale.
    }
    \label{fig:world-model-curves}
\end{figure*}

Figure~\ref{fig:world-model-full} reports paired AULC differences against all
eight baselines, including learned neuron positions, and
Figure~\ref{fig:world-model-curves} shows the corresponding rollout-16 learning
curves. Across the 119 task--baseline pairs for the seven baselines,
LeWi wins 99, ties 6, and loses 14. Including learned neuron positions, it wins
116 of 136 comparisons, ties 6, and loses 14. The per-baseline win--tie--loss
counts are 16--0--1 against Dense, 15--0--2 against $\ell_1$, 13--0--4 against
$\ell_2$, 16--1--0 against $L_0$, 14--1--2 against RigL, 13--3--1 against
Fiedler, 12--1--4 against BIMT, and 17--0--0 against learned neuron positions.

The largest gains occur on low-dimensional control and navigation tasks, while
most ties and losses occur on more complex articulated-control and manipulation
tasks. LeWi also improves on several higher-dimensional and contact-rich
systems, so ambient dimension alone does not explain this pattern. Together
with the CSFRBench results, the world-model results are consistent with LeWi
providing the most benefit when predictive dependencies can be organized into
sparse, low-arity interactions.

\end{document}